\documentclass[final,1p,times,twocolumn]{elsarticle}

\usepackage{amssymb}
\usepackage{amsmath}
\usepackage{caption}
\usepackage{subcaption}
\usepackage{algorithm}
\usepackage{algpseudocode}

\journal{}

\begin{document}

\begin{frontmatter}



\title{Attention Boosted Autoencoder for Building Energy Anomaly Detection}


\author[inst1]{Durga Prasad Pydi}

\affiliation[inst1]{organization={Department of Mechanical Engineering, Indian Institute of Technology Madras },
            city={Chennai},
            postcode={600036}, 
            country={India}}

\author[inst1]{S. Advaith\corref{cor1}}
\cortext[cor1]{Corresponding Author}
\ead{advaith@iitm.ac.in}

\begin{abstract}
Leveraging data collected from smart meters in buildings can aid in developing policies towards energy conservation. Significant energy savings could be realised if deviations in the building operating conditions are detected early, and appropriate measures are taken. Towards this end, machine learning techniques can be used to automate the discovery of these abnormal patterns in the collected data. Current methods in anomaly detection rely on an underlying model to capture the usual or acceptable operating behaviour. In this paper, we propose a novel attention mechanism to model the consumption behaviour of a building and demonstrate the effectiveness of the model in capturing the relations using sample case studies. A real-world dataset is modelled using the proposed architecture, and the results are presented. A visualisation approach towards understanding the relations captured by the model is also presented.
\end{abstract}

\begin{graphicalabstract}
\includegraphics[width=\textwidth]{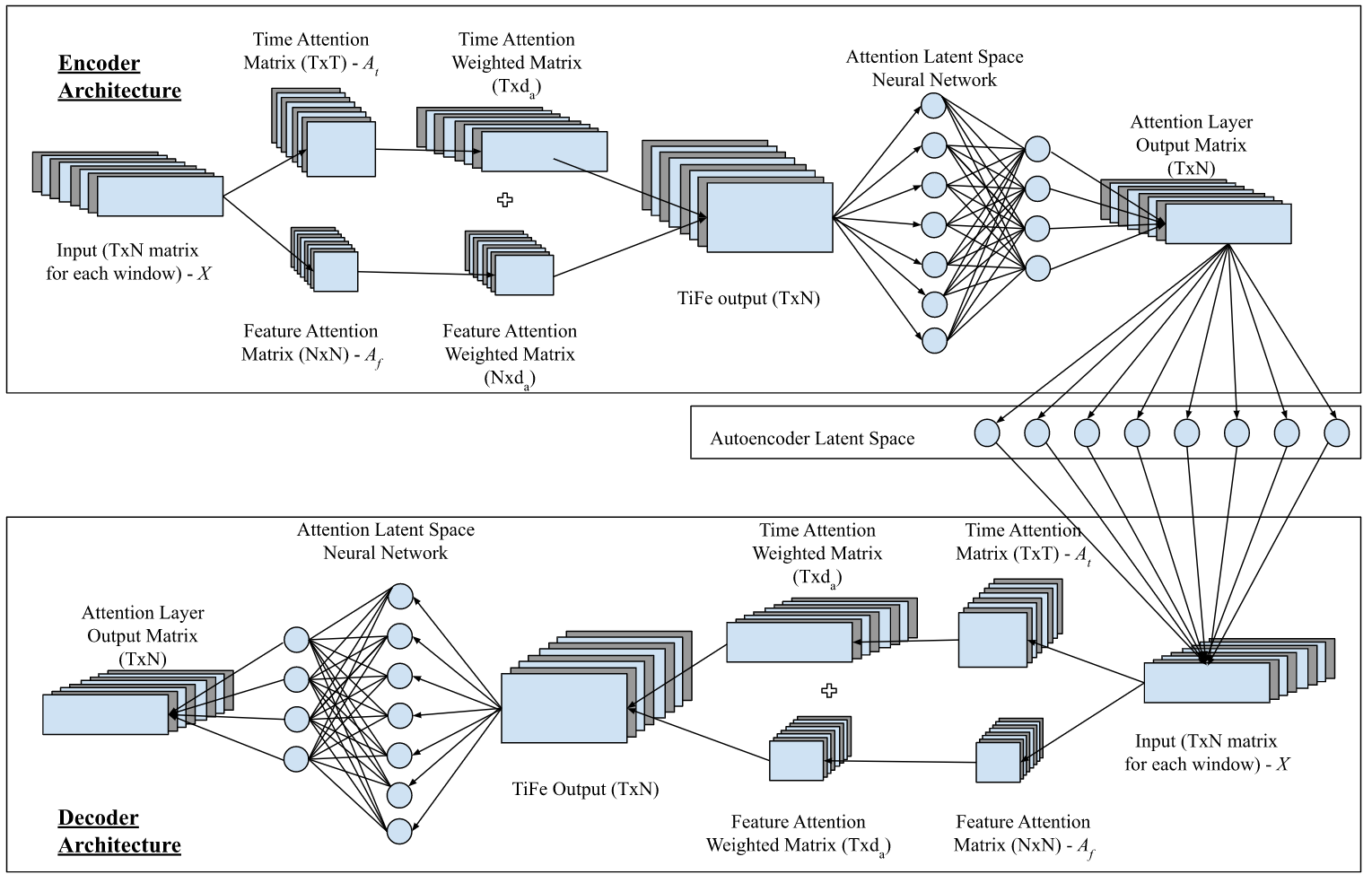}
\end{graphicalabstract}
\begin{highlights}
\item Proposed a novel attention mechanism to capture temporal relations among the features of a multi-variate time series data
\item Provided a qualitative understanding for the anomalous behaviour detected by the model using attention maps
\end{highlights}


\end{frontmatter}


\section{Introduction}
\label{sec:intro}
As per the recent global status report by UNEP\cite{unep_energy}, buildings' energy demand increased by about 4\% from 2020. Operation of buildings accounted for 30\%\cite{iea_report} of global energy consumption with nearly 20\%\cite{iea_cooling} for air-conditioning in buildings and is likely to increase due to increased consumer demand and living standards. Thus reducing operating power consumption without compromising living standards is of utmost importance to achieve ambitious goals towards a sustainable future. Around one-third of this consumption can be attributed to negligent behaviour of consumers, e.g., opening windows with AC turned on or using wrong settings on the ACs\cite{faults}. Automated discovery of such instances has immense potential to provide significant energy savings.  

Anomalies are defined as instances that are relatively rare to occur. The general approach to detect these instances involve modelling the normal behaviour. Events that are not captured by the model will be termed anomalies. Data collected from modern buildings is usually a multivariate time series with multiple features. Manually developing models for such systems becomes cumbersome and is not scalable. Machine learning techniques have demonstrated their ability to capture complex non-linear patterns from the data. These techniques promise to be good candidates for automating the modelling process. A comprehensive review of these techniques applied to building energy anomaly detection is presented in \cite{HIMEUR2021116601}.

Most researchers resorted to developing unsupervised techniques for anomaly detection due to a lack of labelled data \cite{HIMEUR2021116601}. Among these techniques, the recent works include a combination of variational autoencoder\cite{vae} and LSTM\cite{lstm}. Though the techniques produce results superior to the conventional autoregression methods \cite{vae-lstm}, the RNN architecture processes the time frames sequentially, which could potentially increase training times for large-scale multivariate data. Transformer architecture\cite{transformer} alleviates this problem with the attention mechanism. It demonstrated the state of the art results in the field of machine translation. It has been used in \cite{transanomaly} for anomaly detection and demonstrated to be superior to other deep learning techniques. Since the transformer architecture is designed for language processing tasks, it was designed to capture temporal dependencies of various scales. In addition to the temporal dependencies, building energy data has complex interrelationships across the features that need to be considered during the energy consumption modelling. We propose a novel attention mechanism to capture these relationships in a multivariate time series and test the model on synthetically generated and real-world case studies for power consumption in buildings. Specifically, we train our model on the windows generated from the data. The trained model is used to reconstruct the window for test data. The difference between the given window and reconstructed window is labelled as anomaly. In the present work, the model performance in capturing the relations is qualitatively established using attention maps.

In summary, the main contributions are as follows:
\begin{enumerate}
    \item We propose a novel algorithm for anomaly detection in multivariate time series termed as TiFeAuto that can model the complex interrelations among the features across the time windows.
    \item In addition to identifying the anomalous points, we propose to use the attention maps and understand the internal working of the model during window reconstruction, thereby increasing the interpretability of the model.
    \item The model's ability to capture relationships in an unsupervised fashion has been tested on synthetically generated data and a real-world dataset.
\end{enumerate}

\section{Methodology}
\label{sec:method}

\subsection{Problem Formulation}
Consider a multivariate time series $S$ with $N$ features. $S$ is sampled along time to generate windows of length $T$. A matrix denotes each window $X_{o}$ of dimension $T\times N$. Given $M$ such windows, the objective is to develop a model to capture the time-varying relations among the features and reconstruct the given windows. During the training phase, the model extracts relevant features from the data to allow for maximum reconstruction. The model reconstructs the window with the learnt feature relationships during the testing phase. Reconstruction loss is calculated, and a threshold is set to identify points that behave differently than others based on the relations it learnt. In the current work, we propose a combination of novel attention mechanism (referred to as TiFe attention) and an encoder-decoder architecture to reconstruct the samples. Here the TiFe attention serves two purposes:
\begin{itemize}
    \item Increase the robustness of the encoder-decoder architecture to outliers in the training data. The TiFe attention weighs the input features before feeding to encoder-decoder architecture in a way that points not conforming to the relations developed so far during training are given less weightage.
    \item Provides a human-interpretable way to understand the relationships captured by the model. 
\end{itemize}

\subsection{TiFe Attention Model}
\label{sec:TiFe}

Since their introduction, LSTM\cite{lstm} architectures have become a go-to model for time series data. LSTM, being an RNN, is sequential when operating on time windows, leading to significantly longer training times. Also, the hidden state needs sufficient latent space to capture the information of all the previous states. This problem is alleviated using a transformer in \cite{transformer} with an attention mechanism where the model can see information across all the windows and identify any potential relations across the time dimension. This approach allows for parallelisation leading to reduced training time. Like the LSTM model, the transformer architecture constrains the model from capturing all the previous state information in the current state with the same feature dimension. To allow the model to create connections across features of different time steps without losing much information, we propose the below mappings:
\begin{itemize}
    \item \(f:\Re^{T\times N}\ \rightarrow \ \Re^{T\times d_a}\) - To capture relationships across time dimension
    \item \(g:\Re^{N\times T}\ \rightarrow \ \Re^{N\times d_a}\) - To capture relationships across features
    \item \(h:\Re^{(N+T)\times d_a} \rightarrow \ \Re^{T\times N} \) - To utilise the developed mappings above and create a representation of the feature across various time scales
\end{itemize}
where $d_a$ is the latent space dimension of the TiFe Attention (a hyperparameter of the TiFeAE Attention model). Each of these transformations is parameterised with fully connected neural network layers. In addition to reduced training time with parallelisation, the model allows for better interpretation with the help of attention maps.

\begin{figure}[h]
    \centering
    \includegraphics[width=\textwidth]{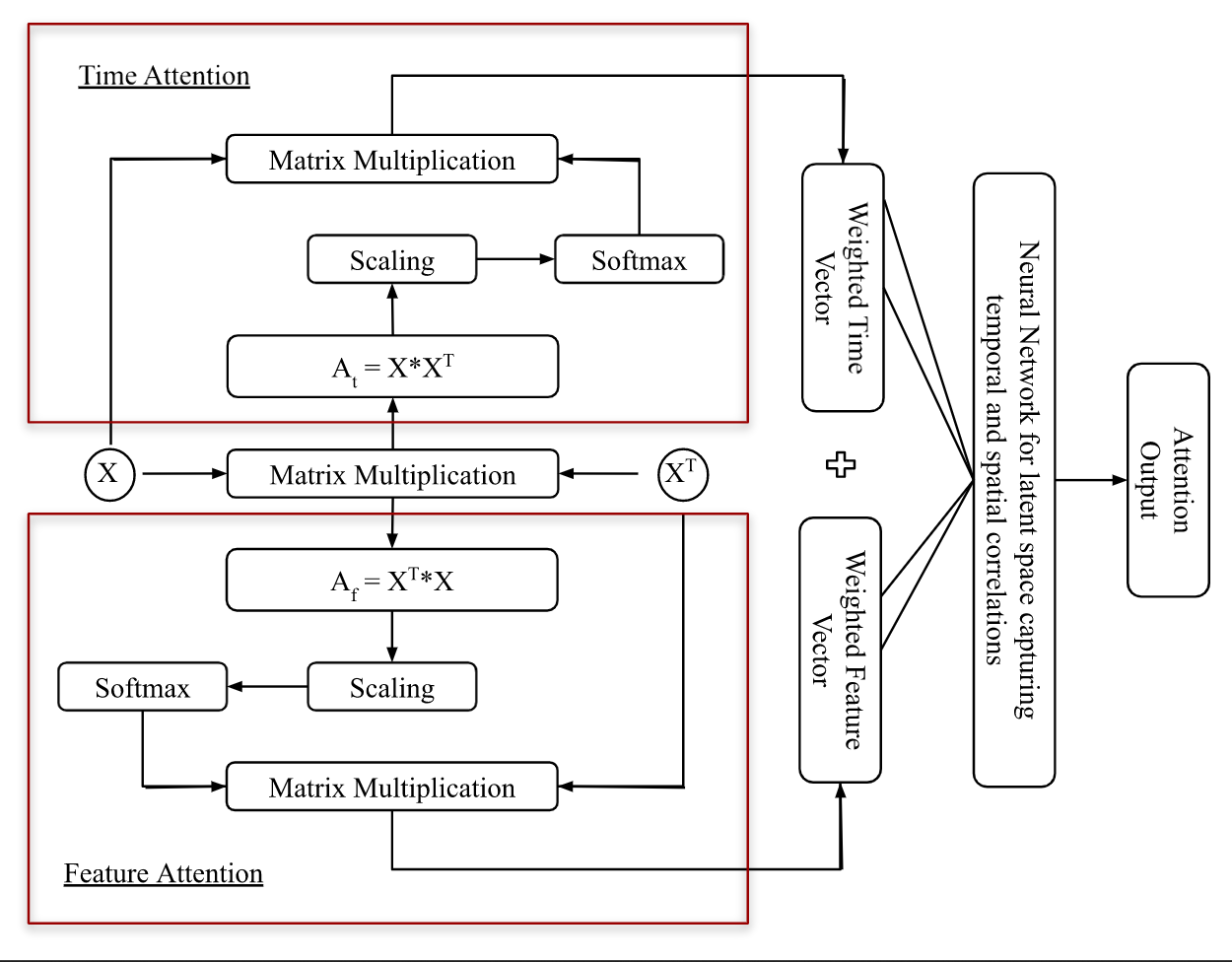}
    \caption{A schematic diagram of TiFeAttention Model Architecture where $X^T$ is the transpose of matrix X and X is $X_{qt}$ or $X_{qf}$ depending on whether time or feature attention is required}
    \label{fig:TiFeAttention}
\end{figure}

Extending from the scaled dot product attention introduced in \cite{transformer}, we define an attention (referred to as TiFe Attention) mechanism that captures relationships across time and feature dimensions as shown in Figure\ref{fig:TiFeAttention}. 

The input windows ($X_o$) are fed to the TiFe attention layers, and the corresponding attention matrices $A_t$, $A_f$ along the time and feature axis are determined by matrix multiplication. The attention matrices are scaled\cite{transformer}, and a softmax function is applied over the matrices to determine the weights of each of the input values along the time/feature dimension. These weights indicate the degree of conformance of the observed data point with the relationships learnt. These weight matrices are multiplied with the original input to obtain reinforced vectors which are fed to a neural network($f_\theta$) that serves as a feature extractor for the auto-encoder described in the Sec\ref{sec:ae}



The outputs of the TiFe Attention Model can be treated as weighted input vectors where a higher value indicates a better conformance of the observed values with the relationships learnt by the model.

\subsection{Encoder-Decoder Model}
\label{sec:ae}

The model has an encoder \( E: \Re^{TxN} \rightarrow \Re^{Txl}\) and decoder
\( E: \Re^{Txl} \rightarrow \Re^{TxN}\) where $l$ is the latent dimension of the model. The encoder attempts to construct a latent space with the given inputs, and the decoder reconstructs from the latent space. Considering the model being trained with usual behaviour data, if the model is fed with an abnormal data point, the reconstruction error would be sufficiently larger. The loss associated with this reconstruction serves as a metric to detect anomalies. Usually, a threshold is set beyond which the points are considered as anomalies. In the current work, we present only a qualitative way f understanding the reconstruction loss with the help of attention maps.

\begin{figure}[!h]
    \centering
    \includegraphics[width=\textwidth]{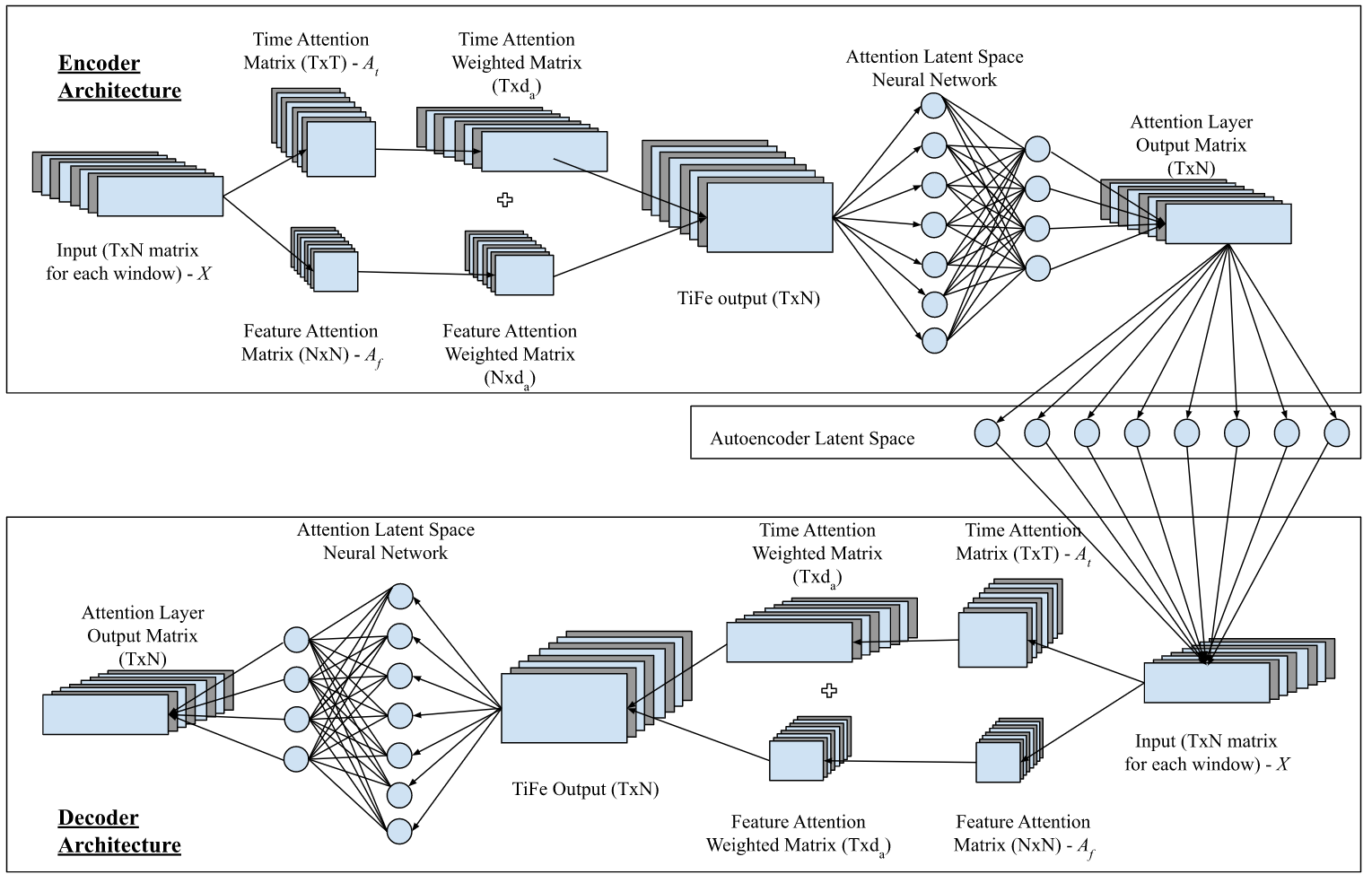}
    \caption{Pictorial representation of the TiFe Attention Model with Encoder-Decoder architecture}
    \label{fig:model}
\end{figure}

The overall architecture is shown in Figure\ref{fig:model}

\subsection{Dataset Preparation}

The paper is organised around three datasets. Two of them are synthetically generated to demonstrate the application of the TiFeAttention model.  A real-world dataset makes up the final one.

\subsubsection{Data-I: Synthetic data for investigating the contributions of TiFe Attention model to the autoencoder}
\label{sec:data1}
\begin{figure}[h]
    \centering
    \includegraphics[width=\textwidth]{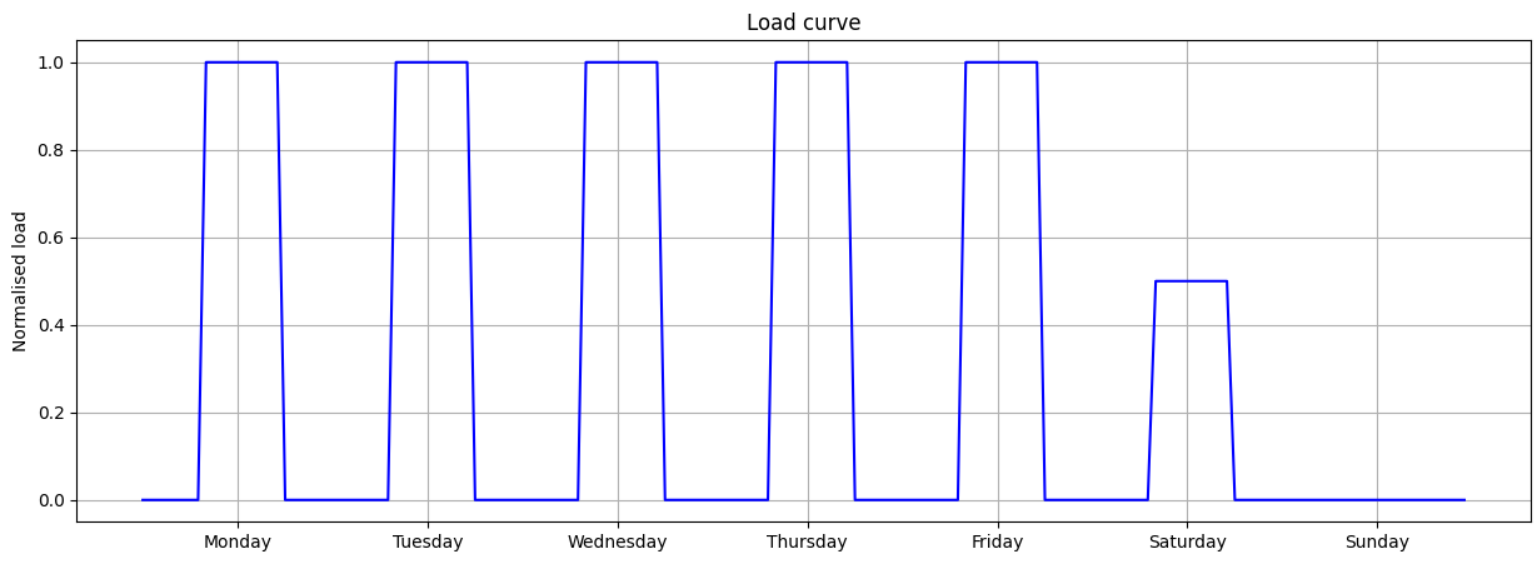}
    \caption{Data-I: Typical hourly power consumption for 1 week}
    \label{fig:data1_norm_prof}
\end{figure}

\begin{figure}[h]
    \centering
    \includegraphics[width=\textwidth]{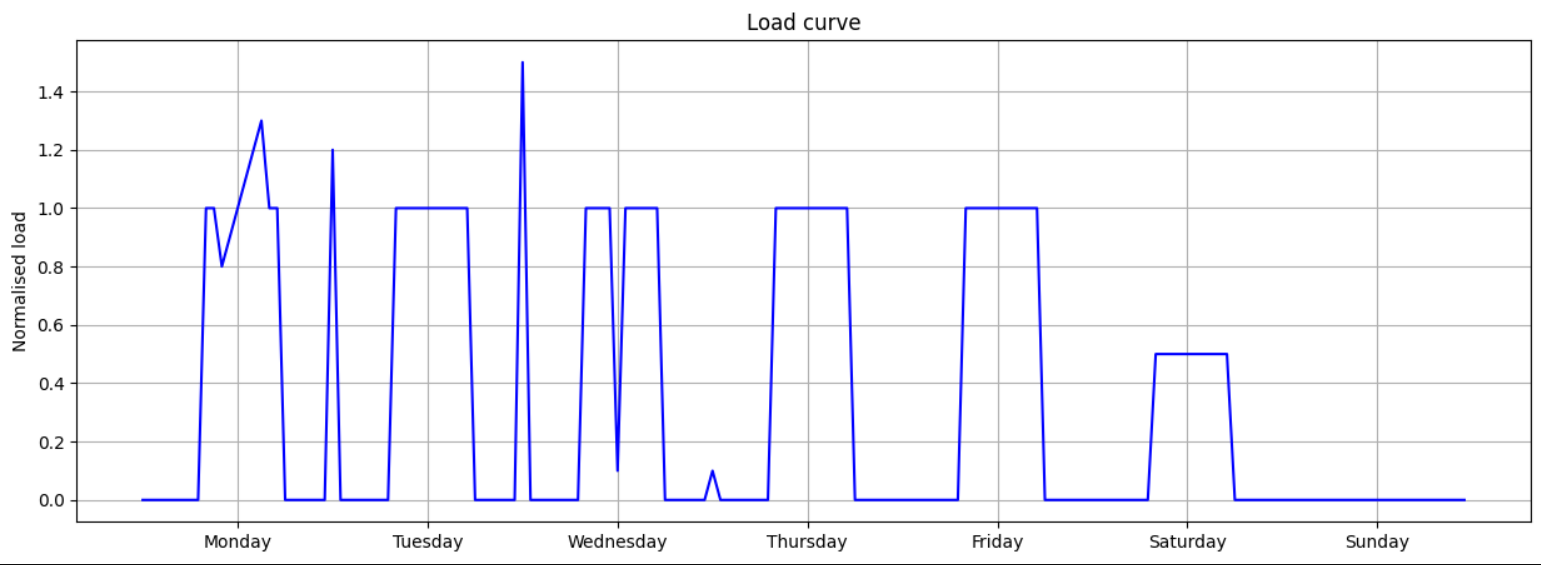}
    \caption{Data-I: Anomalous weekly profile with abnormally high/low value}
    \label{fig:data1_anom_prof}
\end{figure}

The following data have been generated to assess the contribution of the TiFeAttention model to the autoencoder architecture. 
\begin{enumerate}
    \item Algorithm\ref{algo:data1_typ} is used to generate typical weekly load behaviour for 1 year. 
    \item Algorithm\ref{algo:data1_anom} is used to generate anomalous data for 1 week. Data in the third week of the year is replaced with anomalous profile.
    \item An unusual high power consumption at 48 hrs of magnitude 1.5
    \item A low power consumption at 60 hrs of magnitude 0.2 
\end{enumerate}
 Following spikes are also added to the dataset before training the model. If $Y$ is the data obtained after the 
A typical weekly profile alone is shown in Figure\ref{fig:data1_norm_prof}. Artificially introduced anomalies for one week with spikes to resemble anomalies, are shown in Figure\ref{fig:data1_anom_prof}.

\begin{algorithm}
\caption{Algorithm for synthetic data generation for typical profile in Data-I}
\label{algo:data1_typ}
    \begin{algorithmic}
        \Require $N \geq 168$ \Comment(A minimum of 1 week data for training)
        \State $y \gets [0,0,....\text{N times}]$
        \State $i \gets 0$
        \While{$i \leq N$}
            \Comment(Check for day and time (in 24 hrs)
            \If{day is Monday or Tuesday or Wednesday or Thursday or Friday}
                \If{time is between 9 and 19}
                    \State $y[i] \gets 1$
                \EndIf
            \ElsIf{day is Saturday}
                \State $y[i] \gets 0.5$
            \EndIf
        \EndWhile
    \end{algorithmic}  
\end{algorithm}

\begin{algorithm}
\caption{Algorithm for synthetic data generation for 1 week anomalous profile in Data-I}
\label{algo:data1_anom}
    \begin{algorithmic}
        \Require $N \geq 168$ \Comment(A minimum of 1 week data for training)
        \State $y \gets [0,0,....\text{N times}]$
        \State $i \gets 0$
        \While{$i \leq N$}
            \Comment(Check for day and time (in 24 hrs)
            \If{day is Monday or Tuesday or Wednesday or Thursday or Friday}
                \If{time is between 9 and 19}
                    \State $y[i] \gets 1$
                \ElsIf{time is between 0 and 9}
                    \State $y[i] \gets 0.2$
                \EndIf
            \ElsIf{day is Saturday}
                \State $y[i] \gets 0.5$
            \EndIf
        \EndWhile
    \end{algorithmic}  
\end{algorithm}

\subsubsection{Data-II: Synthetic data for investigating the interpretability of the model using time and feature attention maps}
\label{sec:data2}
For illustration purposes, the following data is generated. The test aims to study and understand how the model learnt feature relationships across the variables. It also helps understand how attention maps can be used to tune the model based on consumer requirements.

\begin{figure}[!ht]
    \centering
    \includegraphics[width=\textwidth]{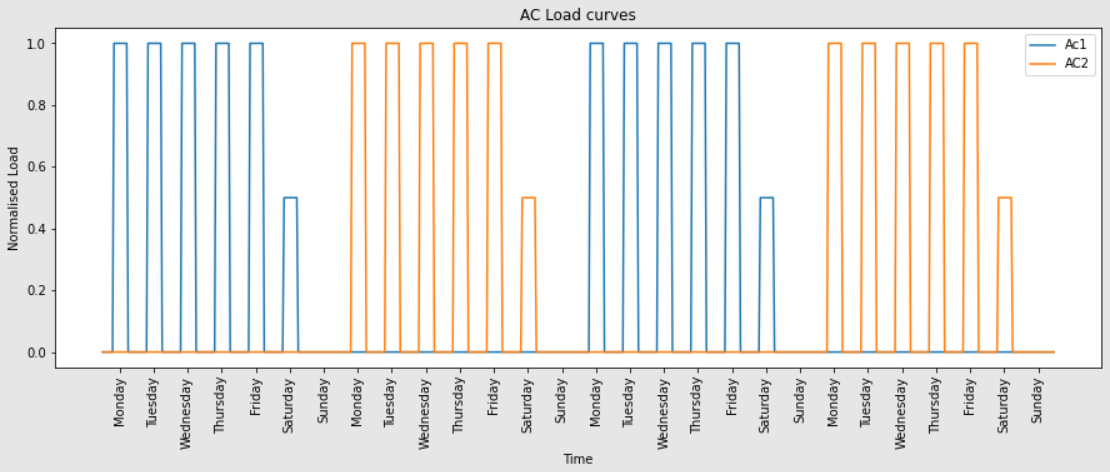}
    \caption{Data-II: Typical hourly power consumption for 4 weeks }
    \label{fig:data2_norm_prof}
\end{figure}

\begin{figure}[!ht]
    \centering
    \includegraphics[width=\textwidth]{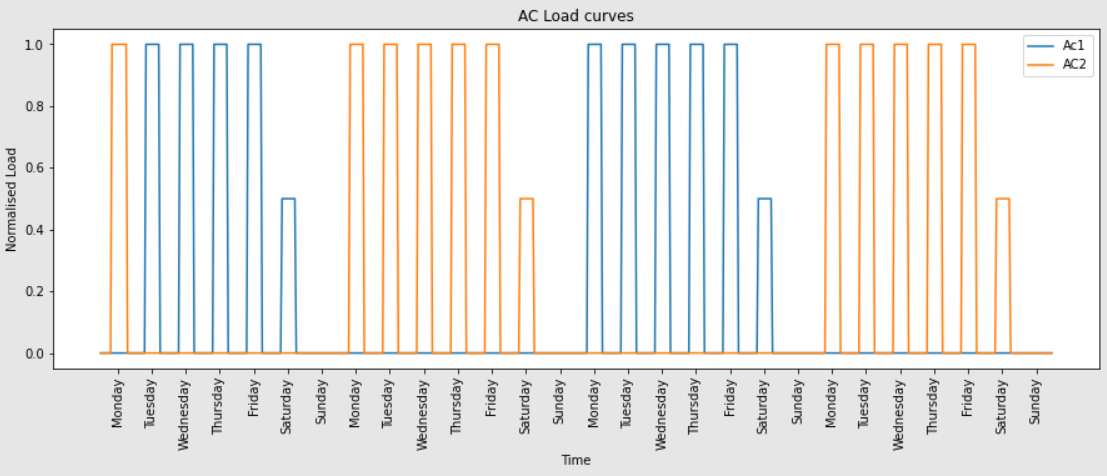}
    \caption{Data-II: Hourly power consumption profile with features interchanged as described in Section\ref{sec:data2}}
    \label{fig:data2_anom_prof}
\end{figure}

Consider a room with two air-conditioning units. In order to meet the room heating load, one of the air-conditioners is sufficient, and from last year's data, it was observed that both operate in a biweekly fashion, as shown in Fig.\ref{fig:data2_norm_prof}.Algorithm\ref{algo:data1_typ} is used to generate data for 1 year with a periodicity of 2 weeks alternating for each air-conditioner. A synthetic anomaly where the operation of 2 air-conditioning units is swapped for one whole day is shown in Figure.\ref{fig:data2_anom_prof}. This data demonstrates how attention maps can be used to understand the relations captured by model in Section\ref{sec:results}.

\subsubsection{Data-III: Real world data of an academic office building}
\label{sec:data3}

The proposed model is tested using a real-world dataset in \cite{cubems}. It has data from over 55 air-conditioning units in an academic office in Thailand sampled every minute. Apart from air-conditioning loads, there are lighting and plug loads. For the current study, only air-conditioning loads shown in Table\ref{tab:data3} are considered. These loads are strategically chosen for the following reasons:
\begin{itemize}
    \item The units belonging to different zones follow different periodic cycles as observed so the model is expected to capture these dependencies feature-wise.
    \item The power consumption of units in the same zone are highly correlated with each other while across different zones are not related in general. The units in the same zone tend to have a load sharing pattern which can only be reconstructed if the model learns to capture the relations among load patterns of units in the same zone at various times. This can be verified by with the attention maps.
\end{itemize}
The missing values are imputed with the mean values and then scaled using Eqn.\ref{eqn:min_max_scaling}.
\begin{equation}
    \label{eqn:min_max_scaling}
    x_i = \frac{x_i-x_{i,min}}{x_{i,max}-x_{i,min}}
\end{equation}
where $i=1,2,...,N$  and $x_{i,min}$,$x_{i,max}$ are the minimum and maximum values of feature $x_i$.

\begin{table}[h]
    \centering
    \begin{tabular}{|c|c|c|}
        \hline
        Floor  & Zone & Number of units \\
        \hline
        1 & 2 & 4 \\
        \hline
        2 & 1 & 1 \\
        \hline
        2 & 2 & 3 \\
        \hline
    \end{tabular}
    \caption{Data considered for model training}
    \label{tab:data3}
\end{table}

A snap of 3 weeks of hourly data for all the 8 AC units is shown in Figure\ref{fig:data3_features}

\begin{figure}[ht]
    \centering
    \begin{subfigure}[b]{\textwidth}
        \includegraphics[width=\textwidth]{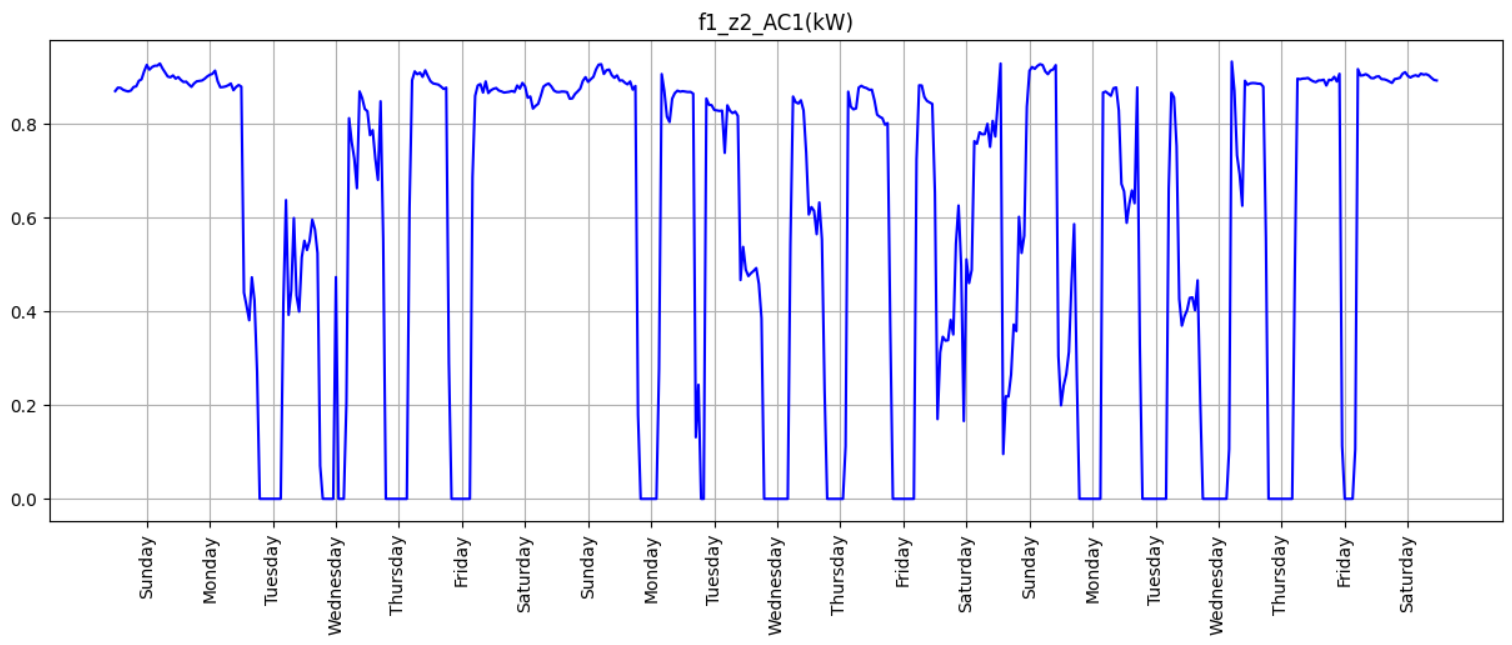}
        \caption{Data-III: Typical hourly consumption profile for AC in Floor1, Zone 2, AC 1}
        \label{fig:data3_feat1}
    \end{subfigure}
    \begin{subfigure}[b]{\textwidth}
        \includegraphics[width=\textwidth]{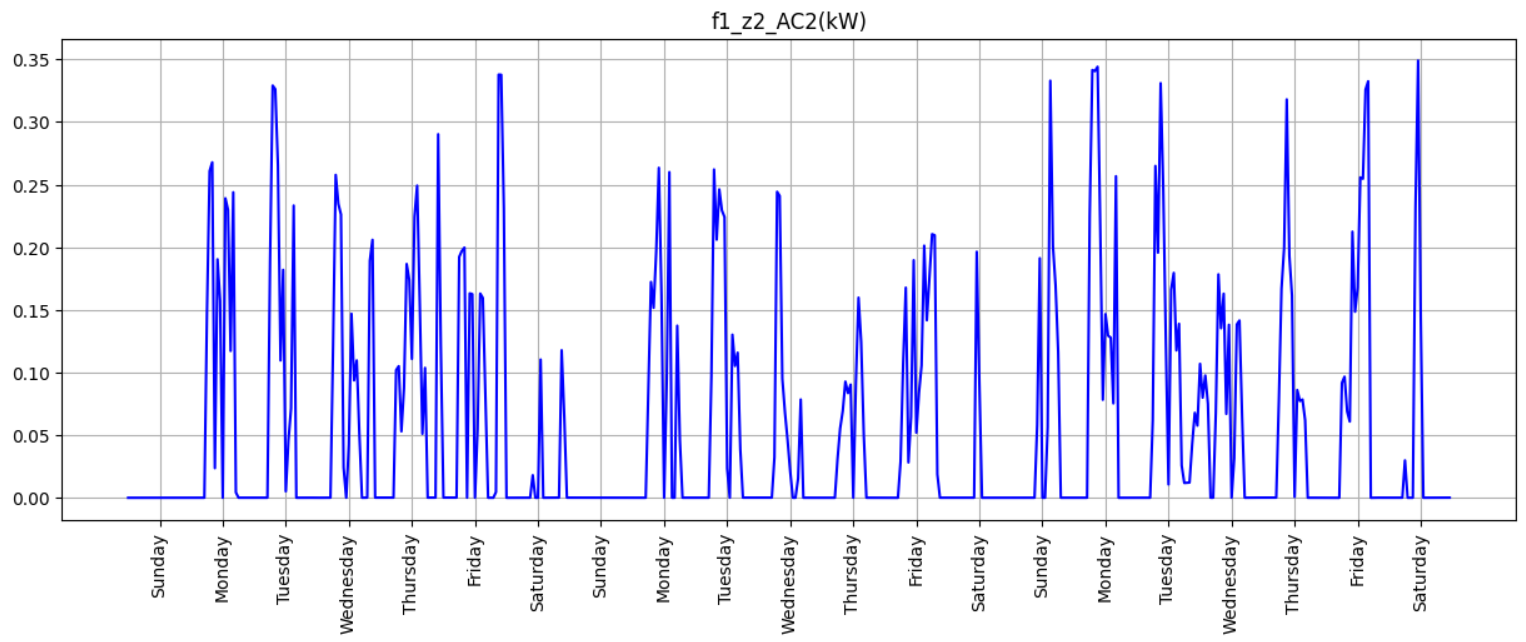}
        \caption{Data-III: Typical hourly consumption profile for AC in Floor1, Zone 2, AC 2}
        \label{fig:data3_feat2}
    \end{subfigure}
    \begin{subfigure}[b]{\textwidth}
        \includegraphics[width=\textwidth]{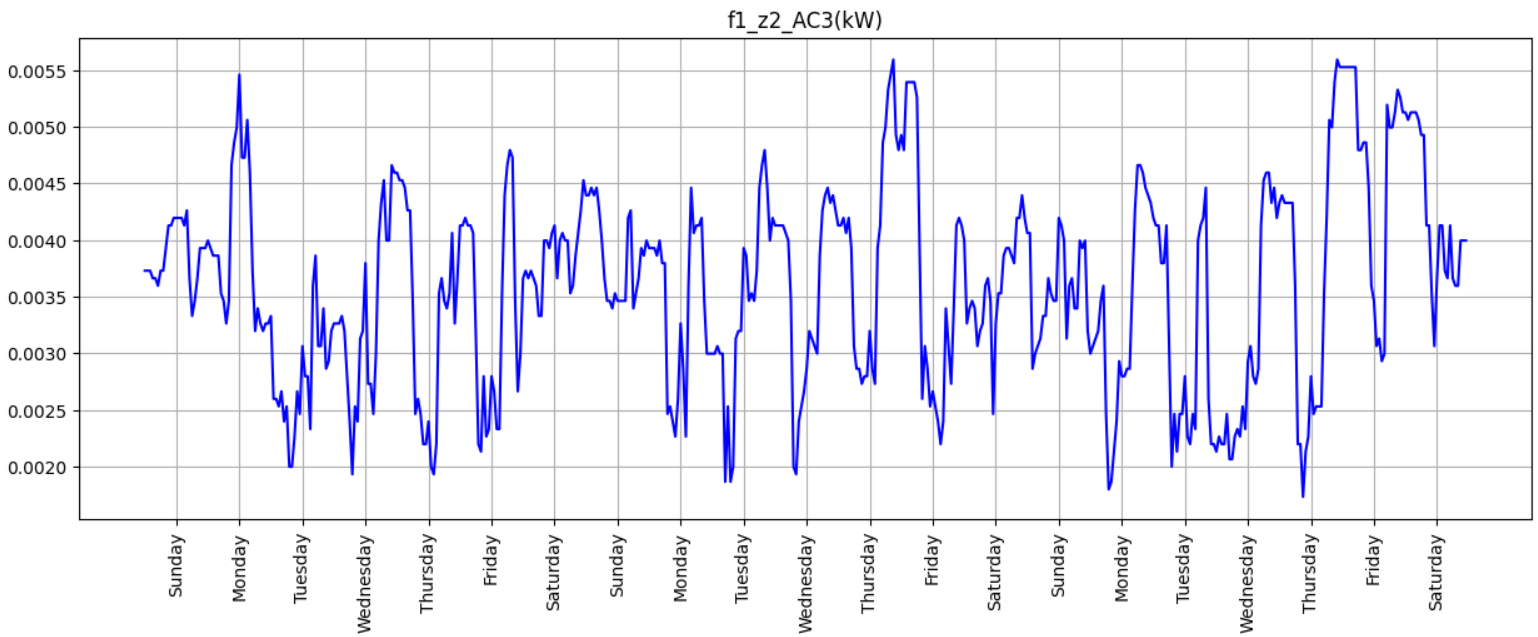}
        \caption{Data-III: Typical hourly consumption profile for AC in Floor1, Zone 2, AC 3}
        \label{fig:data3_feat3}
    \end{subfigure}
    \caption{Data-III}
\end{figure}
\begin{figure}[ht]\ContinuedFloat
    \centering
    \begin{subfigure}[b]{\textwidth}
        \includegraphics[width=\textwidth]{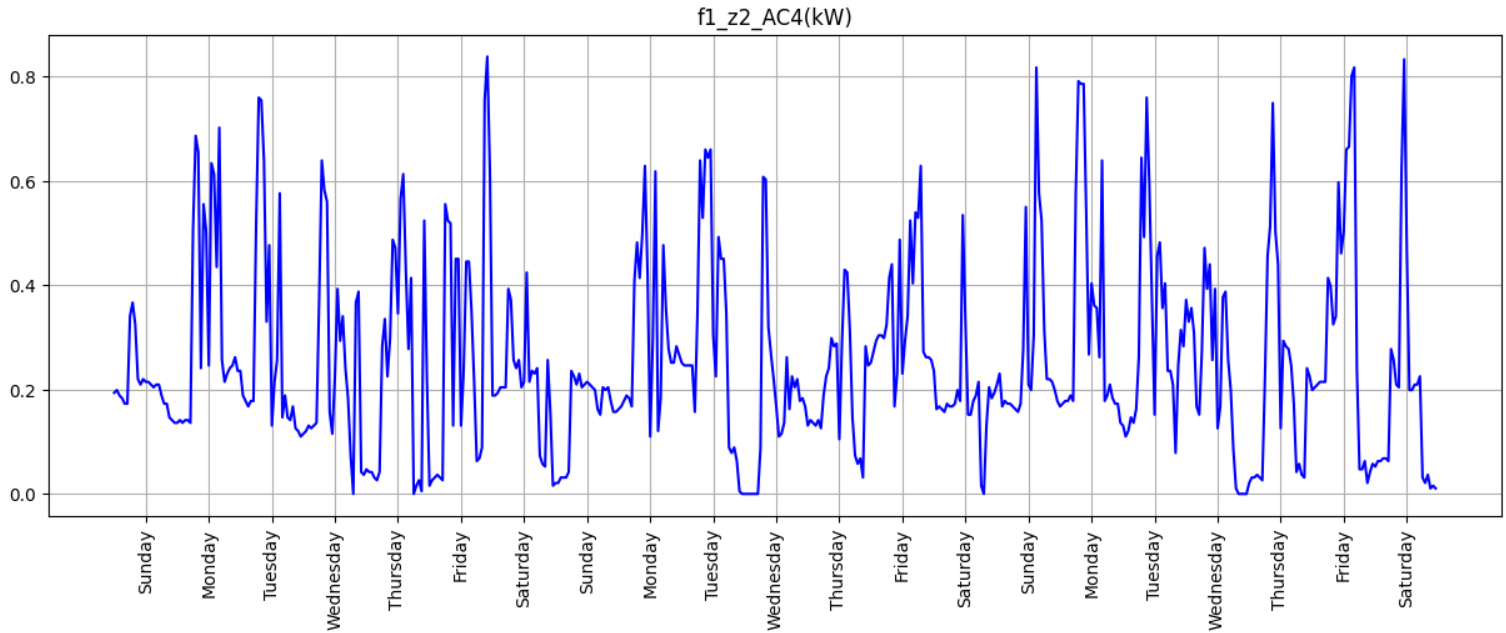}
        \caption{Data-III: Typical hourly consumption profile for AC in Floor1, Zone 2, AC 4}
        \label{fig:data3_feat4}
    \end{subfigure}
    \begin{subfigure}[b]{\textwidth}
        \includegraphics[width=\textwidth]{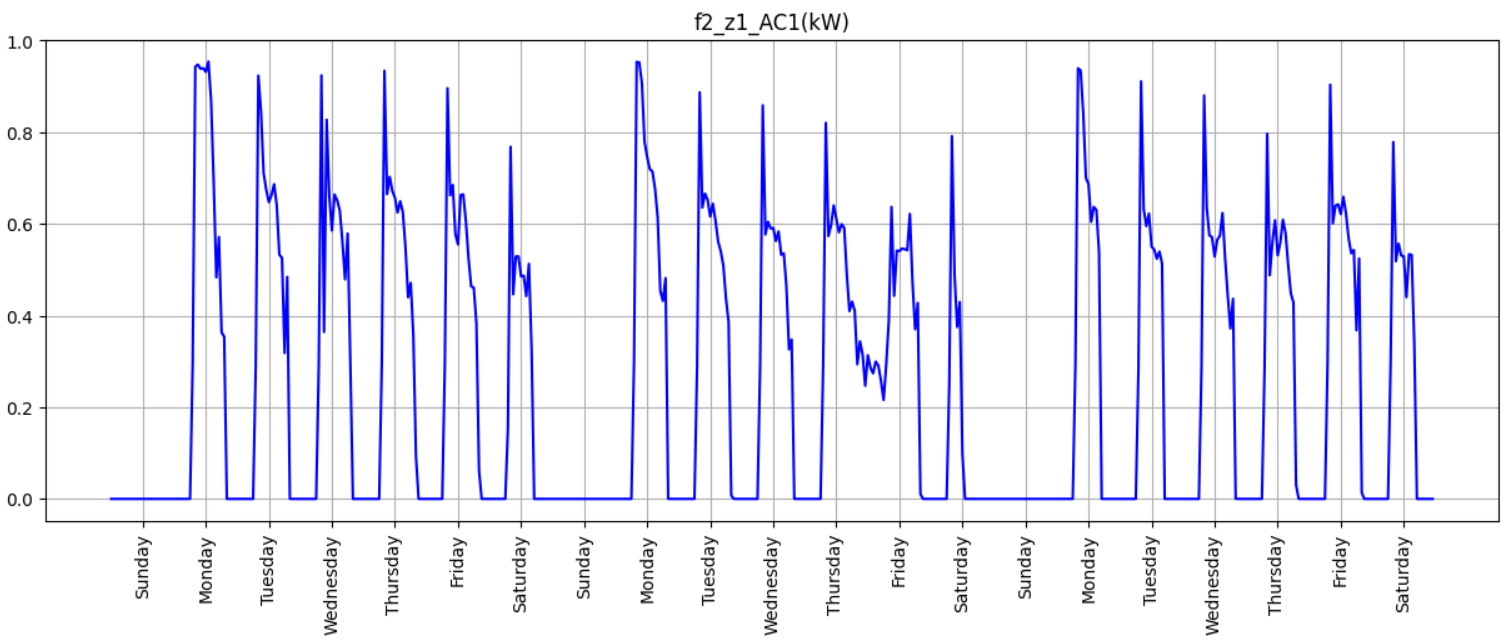}
        \caption{Data-III: Typical hourly consumption profile for AC in Floor2, Zone 1, AC 1}
        \label{fig:data3_feat5}
    \end{subfigure}
    \begin{subfigure}[b]{\textwidth}
        \includegraphics[width=\textwidth]{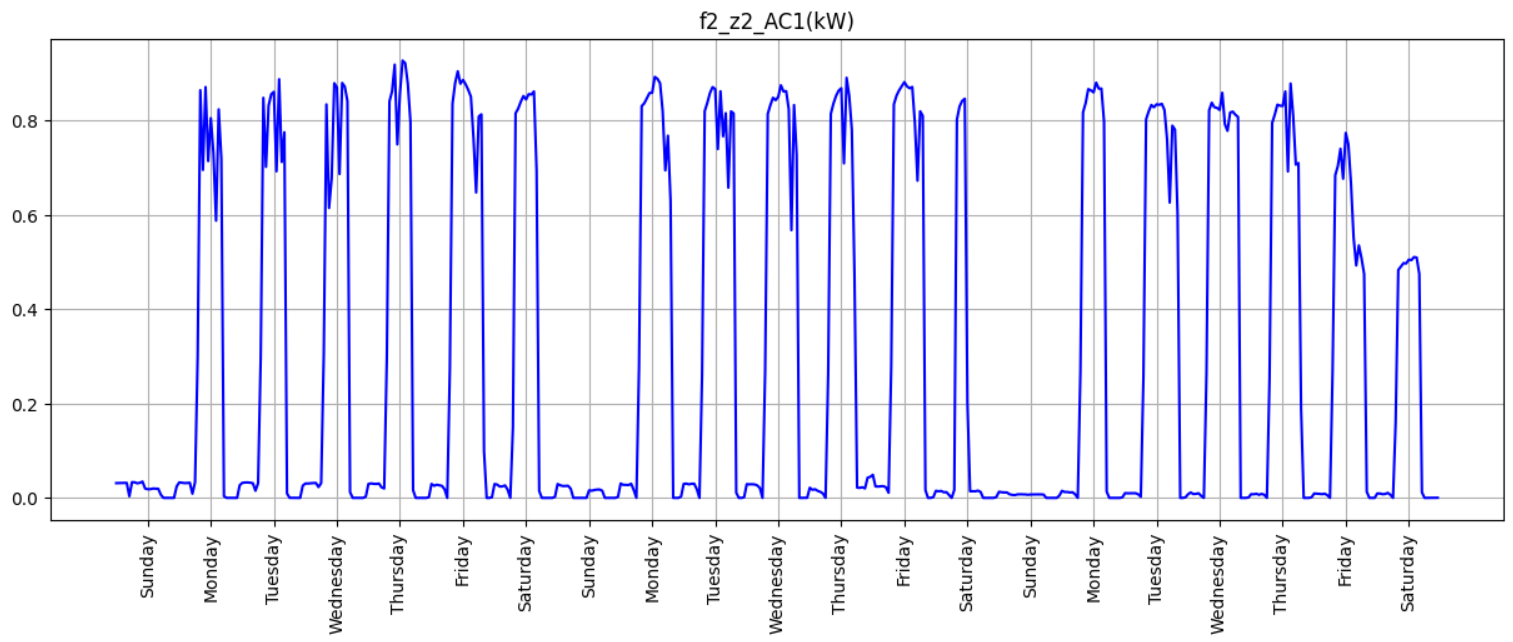}
        \caption{Data-III: Typical hourly consumption profile for AC in Floor2, Zone 2, AC 1}
        \label{fig:data3_feat6}
    \end{subfigure}
    \caption{Data-III (contd.)}
\end{figure}
\begin{figure}[ht]\ContinuedFloat
    \begin{subfigure}[b]{\textwidth}
        \includegraphics[width=\textwidth]{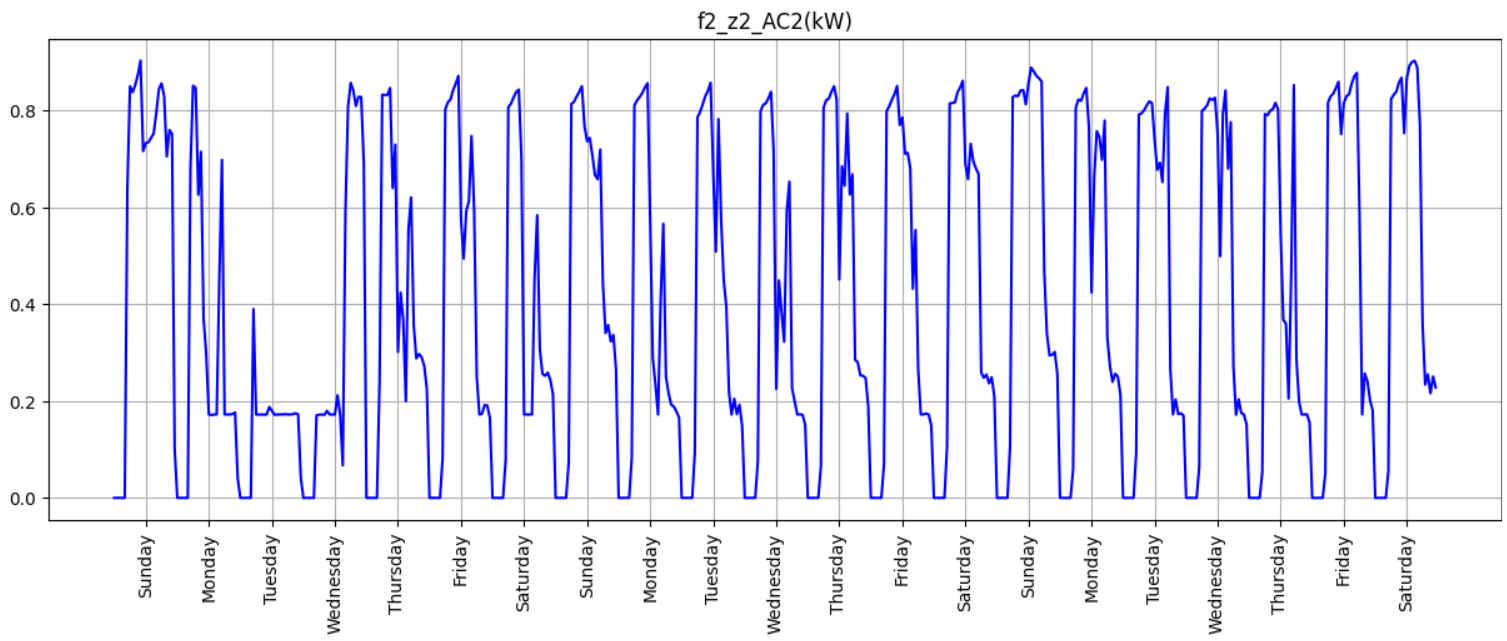}
        \caption{Data-III: Typical hourly consumption profile for AC in Floor2, Zone 2, AC 2}
        \label{fig:data3_feat7}
    \end{subfigure}
    \begin{subfigure}[b]{\textwidth}
        \includegraphics[width=\textwidth]{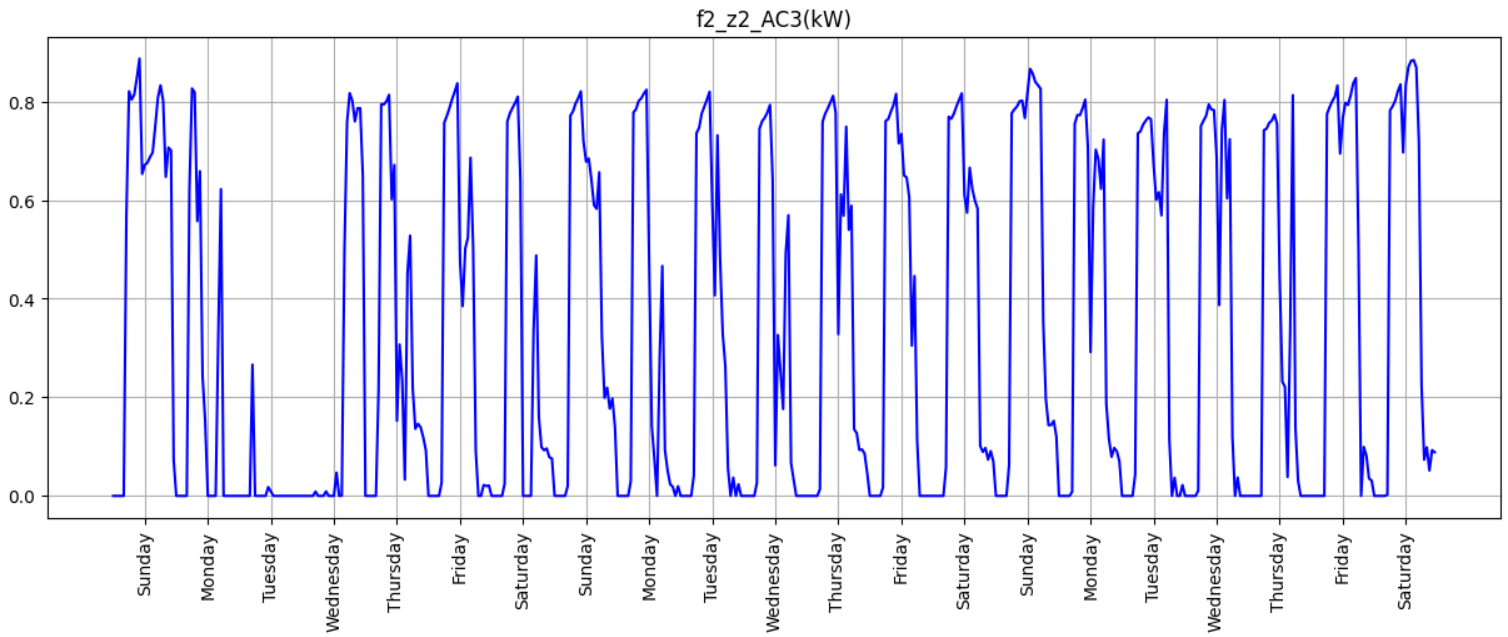}
        \caption{Data-III: Typical hourly consumption profile for AC in Floor2, Zone 2, AC 3}
        \label{fig:data3_feat8}
    \end{subfigure}
    \caption{Data-III (contd.)}
    \label{fig:data3_features}
\end{figure}

\subsection{Model training}
\label{sec:training}
Hourly rate samples are used throughout the study. Adam optimizer with default parameters is used in tensorflow for all the models. The hyperparameters of the models are tabulated in Table\ref{tab:model_parameters}
\begin{table}[h]
    \centering
    \caption{Model Parameters}
    \begin{tabular}{|c|c|c|c|}
        \hline
        Parameter & Data-I & Data-II & Data-III \\
        \hline
        $d_a$ & 64 & 64 & 32\\
        \hline
        $l$ & 64 & 32 & 16 \\
        \hline
        batch size & 128 & 256 & 128 \\
        \hline
        window length & 168 & 336 & 168 \\
        \hline
        number of features & 1 & 2 & 8\\
        \hline
    \end{tabular}
    \label{tab:model_parameters}
\end{table}

\section{Results and Discussions}
\label{sec:results}

\subsection{Data-I: Investigating the contributions of TiFe Attention model to the autoencoder}
\label{sec:results1}

\begin{figure}[!h]
    \centering
    \begin{subfigure}{0.45\textwidth}
        \includegraphics[width=\textwidth]{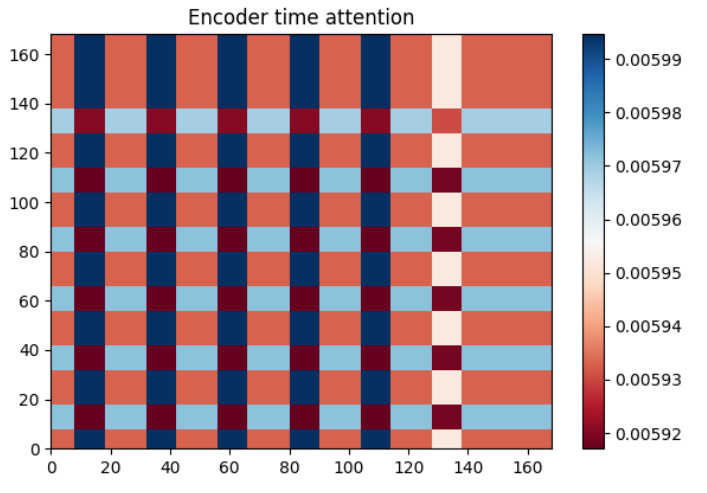}
        \caption{Attention map for usual pattern}
        \label{fig:data1_norm_time_attn}
    \end{subfigure}
    \begin{subfigure}{0.45\textwidth}
        \includegraphics[width=\textwidth]{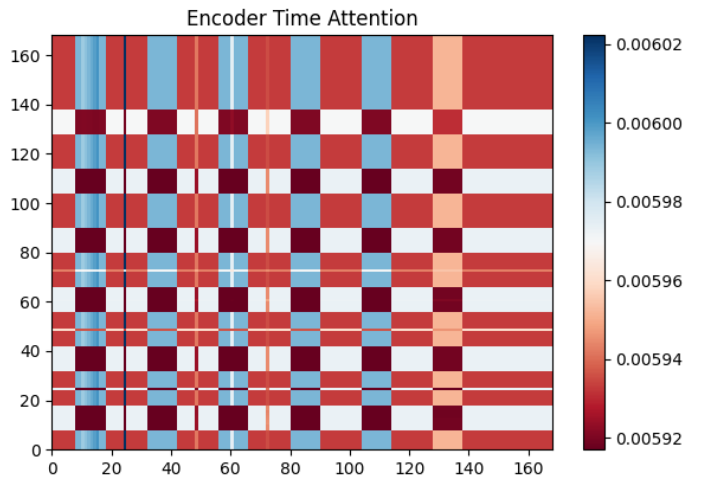}
        \caption{Attention map for window with anomalous spike}
        \label{fig:data1_anom_time_attn}
    \end{subfigure}
    \caption{Data-I: Comparison of attention maps for normal and anomalous patterns}
    \label{fig:data1_time_attn}
\end{figure}

\begin{figure}[!ht]
    \centering
    \begin{subfigure}[b]{\textwidth}
        \includegraphics[width=\textwidth]{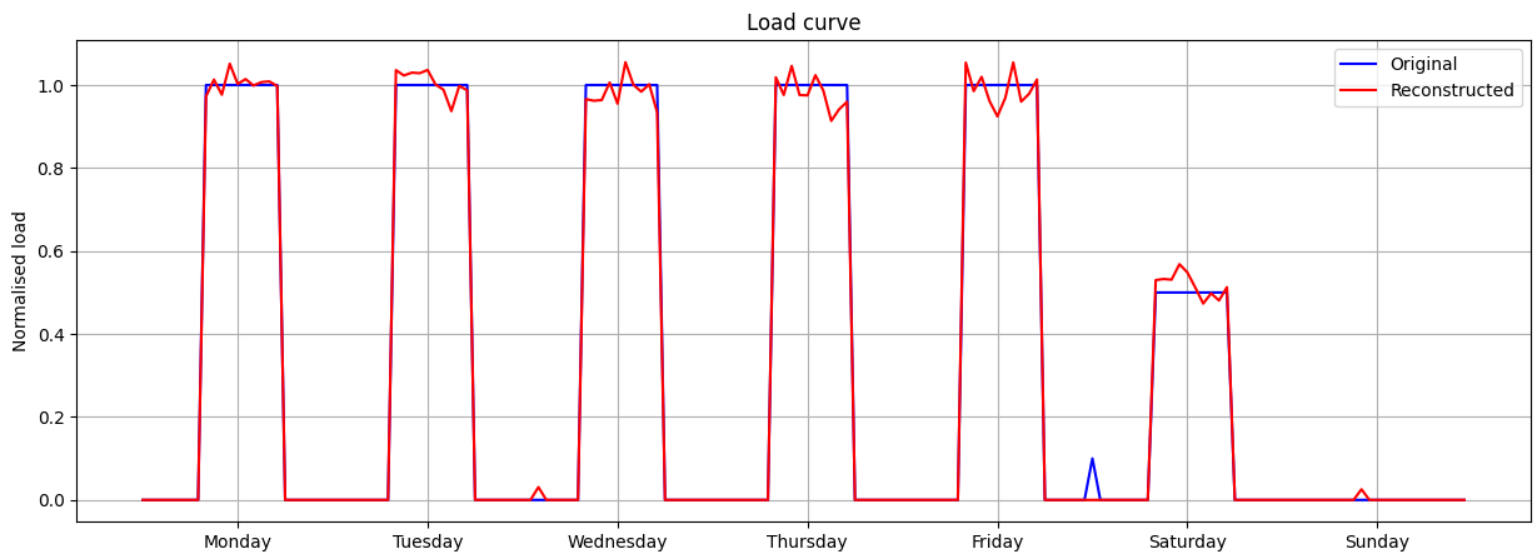}
        \caption{Normal reconstruction profile with TiFe Attention}
        \label{fig:data1_norm_reco}
    \end{subfigure}
    \begin{subfigure}[b]{\textwidth}
        \includegraphics[width=\textwidth]{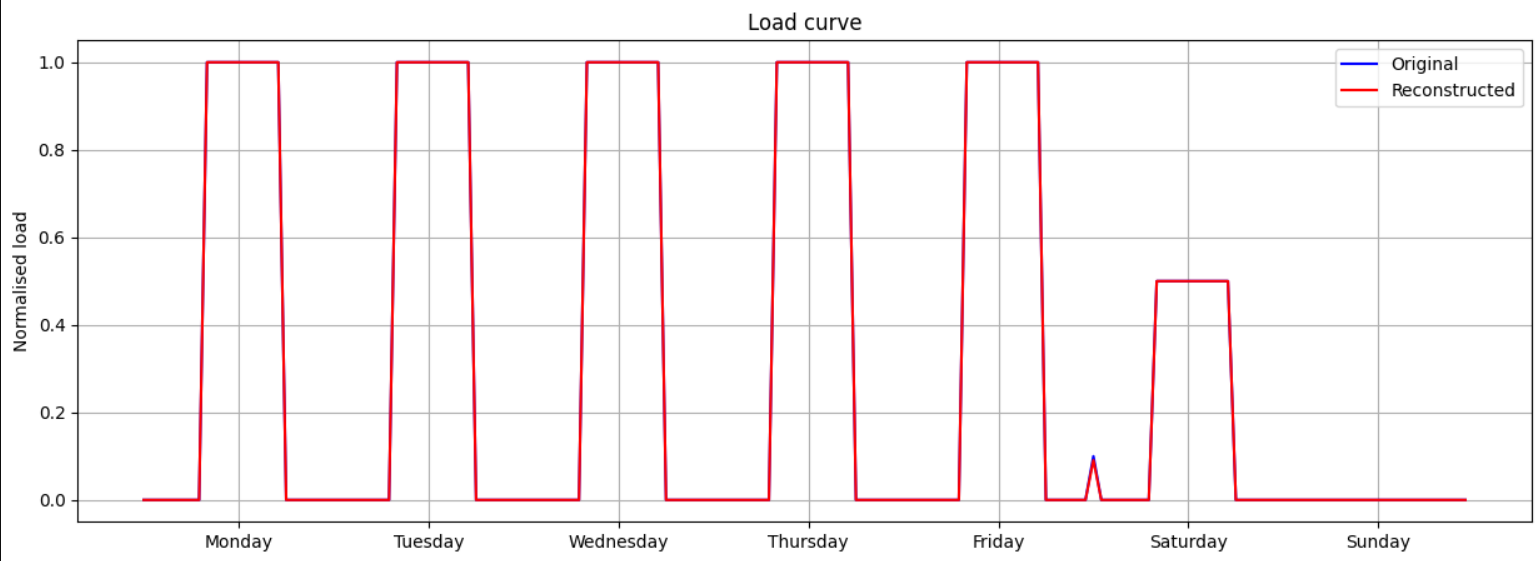}
        \caption{Normal reconstruction profile without TiFe Attention}
        \label{fig:data1_norm_reco_ae}
    \end{subfigure}
    \caption{Data-I: Comparison of reconstruction profiles with and without TiFe attention}
    \label{fig:data1_comp_prof}
\end{figure}

\begin{figure}[!ht]
    \centering
    \begin{subfigure}[b]{\textwidth}
        \includegraphics[width=\textwidth]{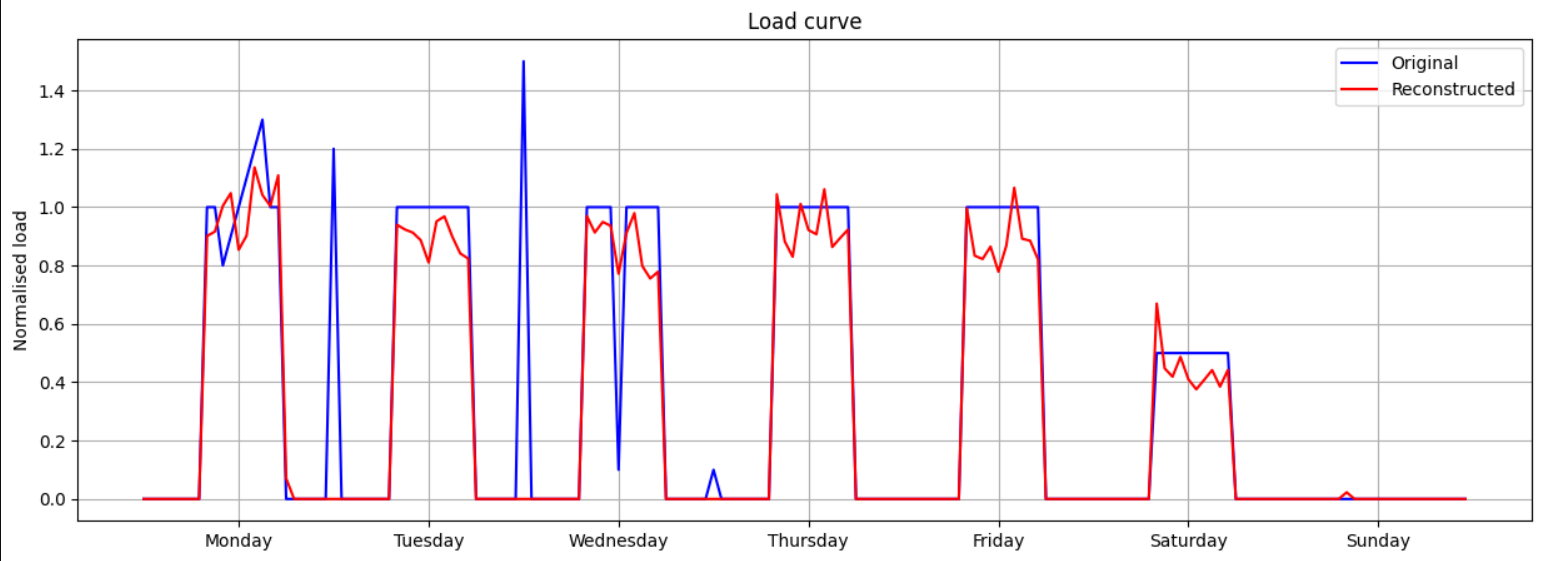}
        \caption{Normal reconstruction profile with TiFe Attention}
        \label{fig:data1_anom_reco}
    \end{subfigure}
    \begin{subfigure}[b]{\textwidth}
        \includegraphics[width=\textwidth]{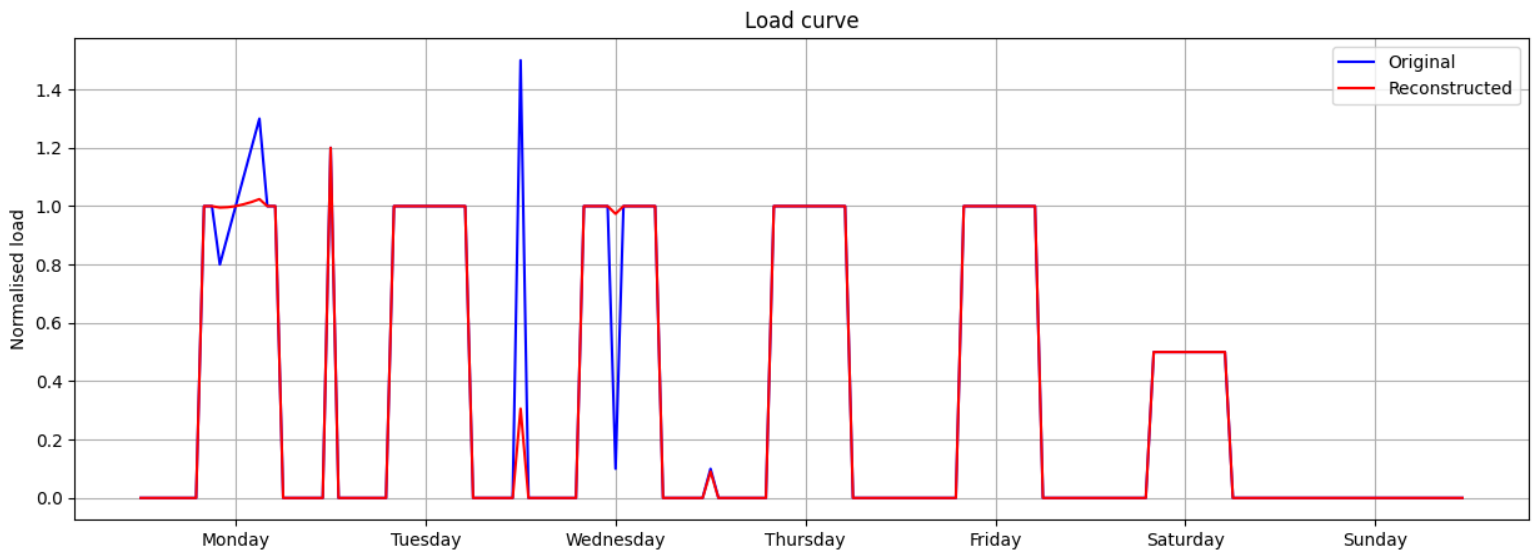}
        \caption{Normal reconstruction profile without TiFe Attention}
        \label{fig:data1_anom_reco_ae}
    \end{subfigure}
    \caption{Data-I: Comparison of reconstruction profiles with and without TiFe attention}
    \label{fig:data1_comp_prof_anom}
\end{figure}

To demonstrate the contributions of TiFe Attention model, data as described in Section\ref{sec:data1}. The model is trained on the data with anomalies using the proposed model and an auto encoder without the TiFe Attention model but of same latent size. A comparison of the reconstruction profiles in windows of usual behaviour is shown in Figure\ref{fig:data1_comp_prof}. Both the models were able to capture the original distribution. A comparison of the windows with abnormal patterns is shown in Figure\ref{fig:data1_comp_prof_anom}. Though the autoencoder alone is able to reconstruct the observed window, it lacked contextual information to constrain it from reconstructing the peak observed on early morning of Sunday in Figure\ref{fig:data1_anom_reco_ae}. TiFe Attention model aided in providing the contextual information by reducing the weight of the sample for reconstruction as visualised in the attention map shown in Figure\ref{fig:data1_time_attn}. It can be observed in Figure\ref{fig:data1_anom_time_attn}, the bands are lightened in regions with unusual spikes indicating the presence of anomalies reinforcing that TiFe Attention model provided contextual information to the encoders.

\subsection{Data-II: Investigating the interpretability of the model using time and feature attention maps}
\label{sec:results2}

\begin{figure}[ht]
    \centering
    \begin{subfigure}[b]{\textwidth}
        \includegraphics[width=\textwidth]{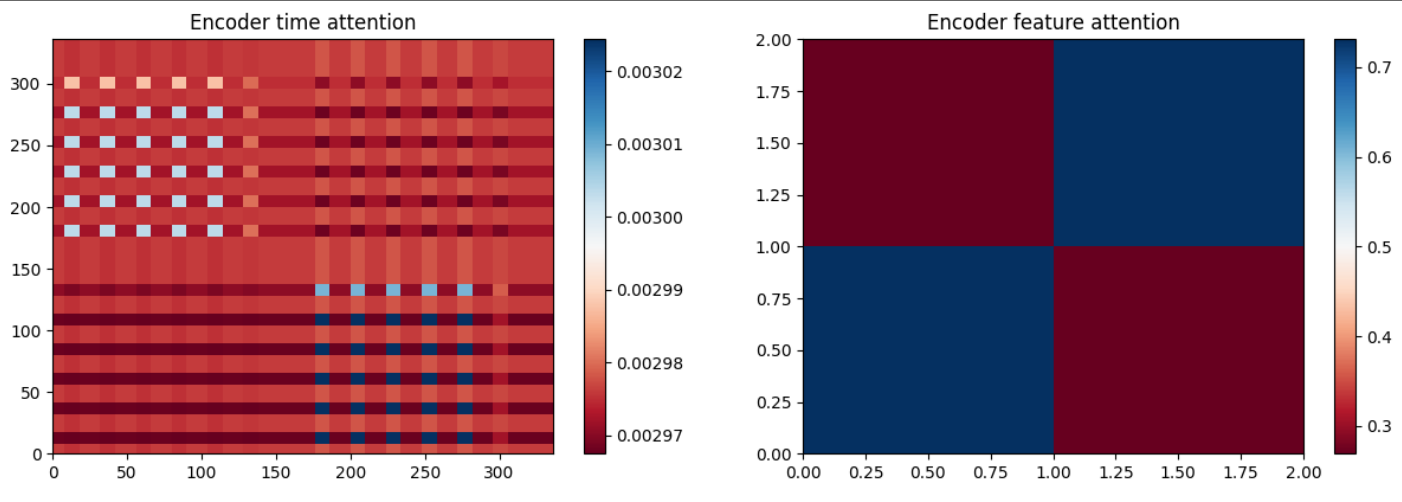}
        \caption{Attention map for normal usage}
        \label{fig:data2_norm_attn}
    \end{subfigure}
    \begin{subfigure}[b]{\textwidth}
        \includegraphics[width=\textwidth]{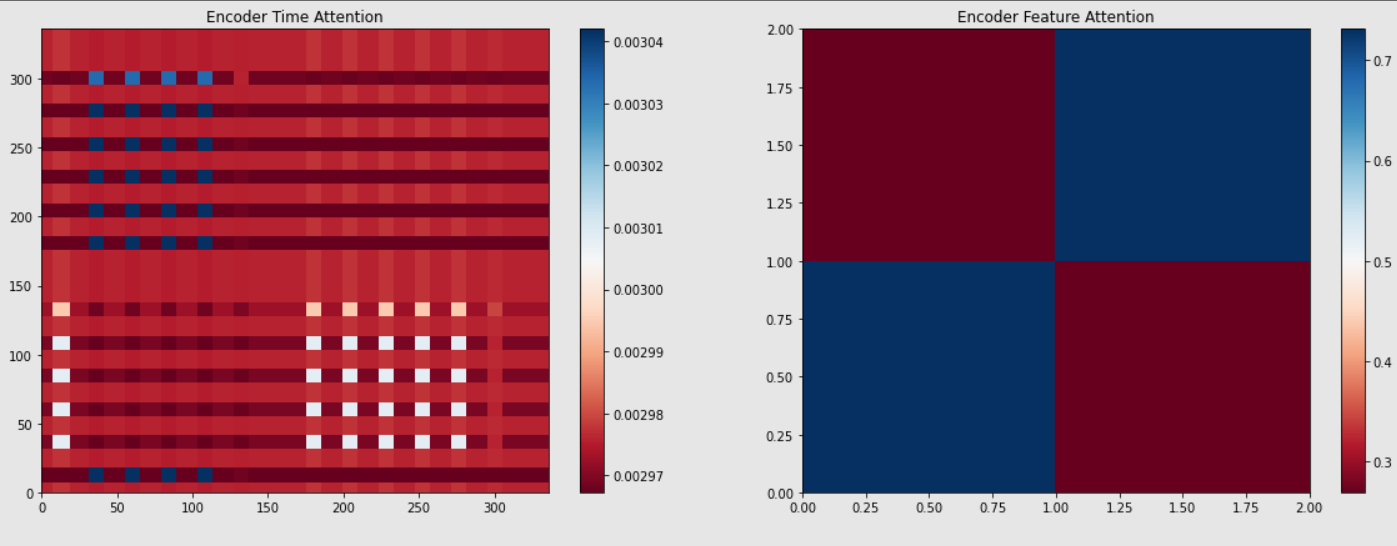}
        \caption{Attention map for anomalous usage}
        \label{fig:data2_anom_attn}
    \end{subfigure}
    \caption{Data-II: Comparison of attention maps }
    \label{fig:data2_comp_attn}
\end{figure}

The model is trained on Data-II (described in Section\ref{sec:data2} and the obtained attention maps are shown in Figure\ref{fig:data2_comp_attn}. As seen in Figure\ref{fig:data2_anom_prof} vs Figure\ref{fig:data2_norm_prof}, the only difference exists on Monday when AC2 operates instead of AC1. The unusually low value for AC1 triggered the TiFe Attention model to increase the weights for the observed values for reconstruction as seen in Figure\ref{fig:data2_anom_attn} where a patch with higher weights are found for the first week. Also both normal and anomalous feature attention maps are similar which indicate that the values being observed do not indicate any abnormality but only the sequence in which they appear is an anomaly. These results signify the importance of attention maps in understanding whether relationships captured and used by the model for reconstructing the sequence are valid. They also help in understanding if the reconstruction error is due to under fitting of the model or an underlying anomaly.

\subsection{Data-III: Investigating model performance on a real world data }
\begin{figure}[ht]
    \centering
    \begin{subfigure}[b]{\textwidth}
        \includegraphics[width=\textwidth]{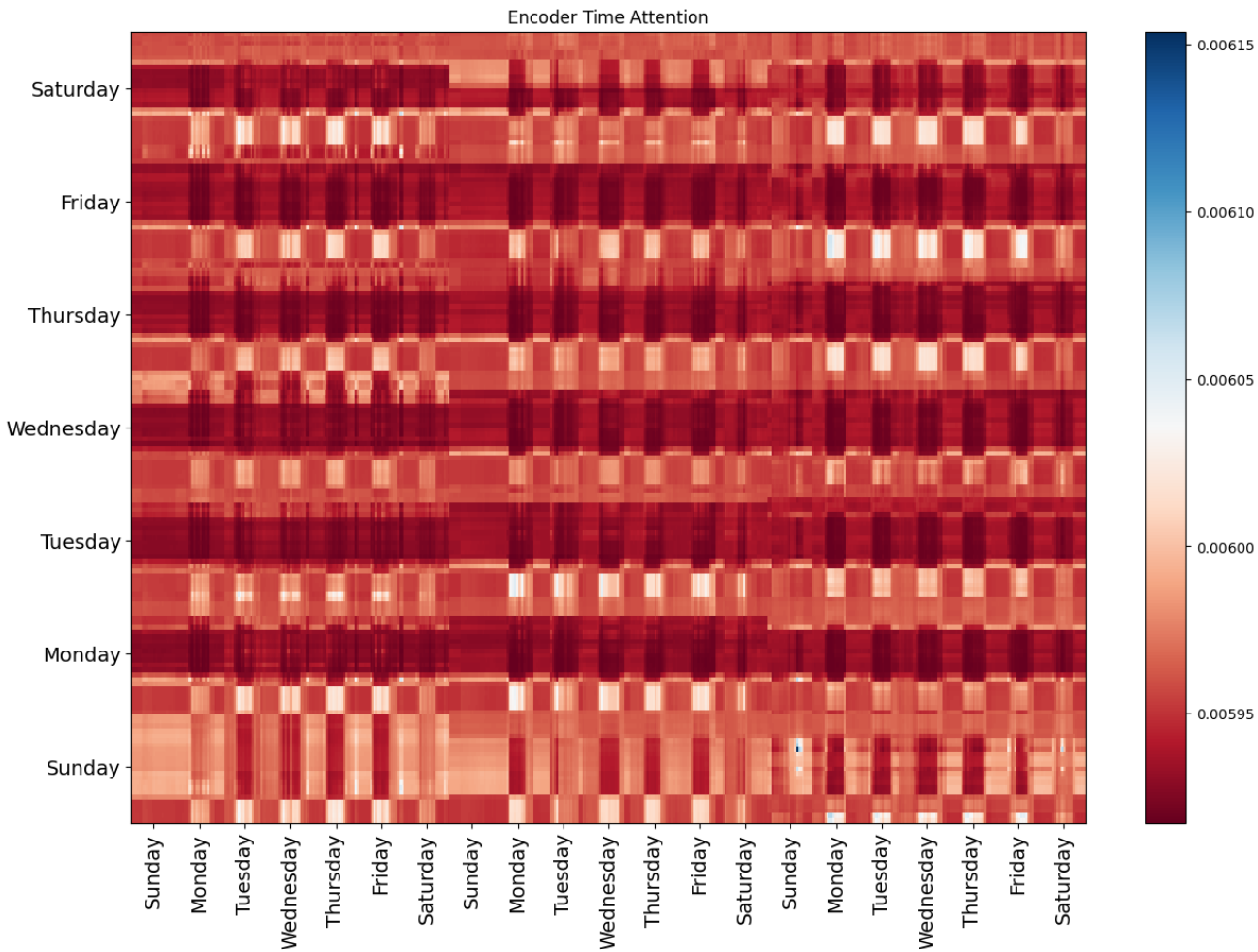}
        \caption{Encoder Time Attention Map}
        \label{fig:data3_time_attn}
    \end{subfigure}
    \begin{subfigure}[b]{\textwidth}
        \includegraphics[width=\textwidth]{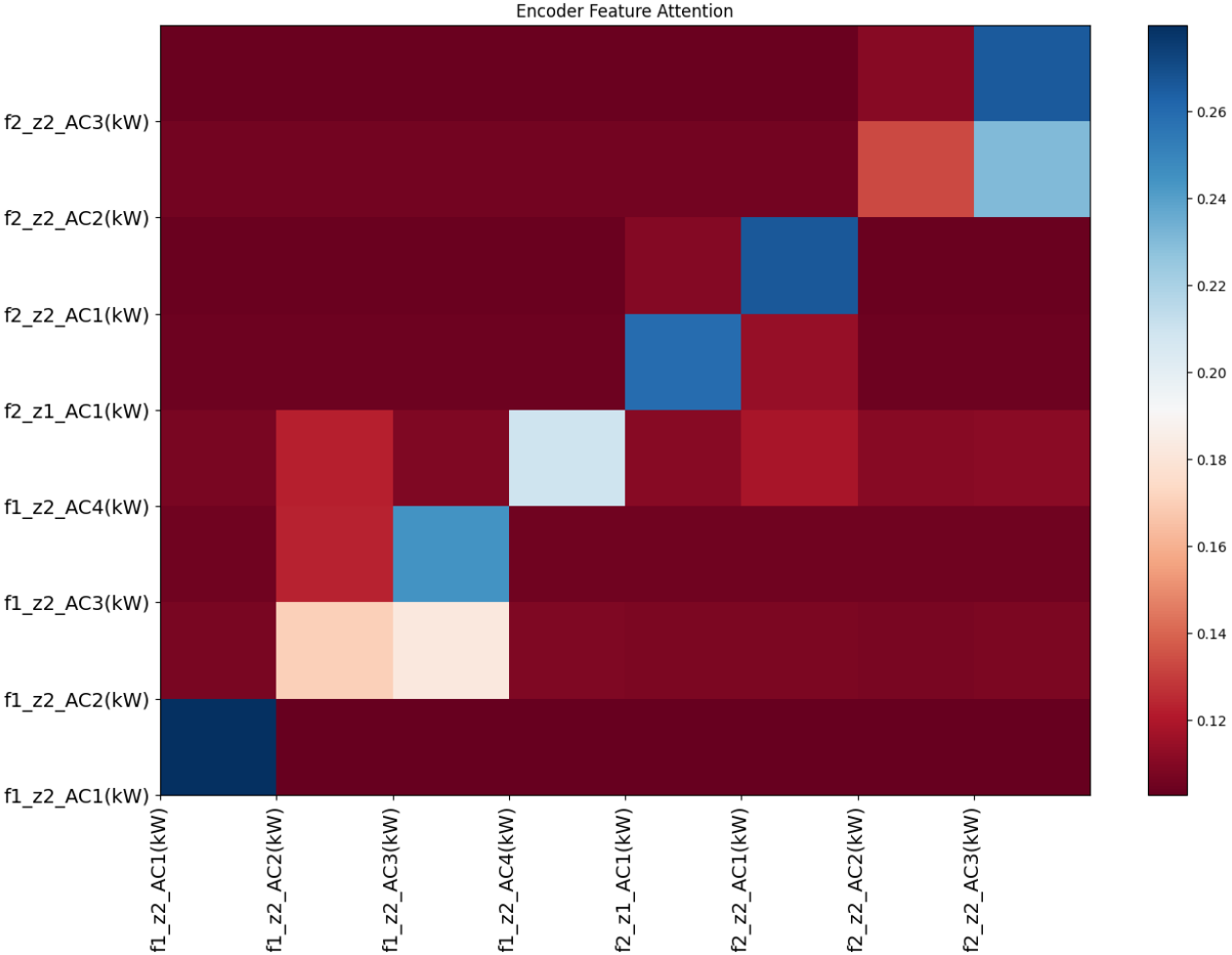}
        \caption{Encoder Feature Attention Map}
        \label{fig:data3_feat_attn}
    \end{subfigure}
    \caption{Data-III: Attention Maps from TiFe Attention model}
    \label{fig:data3_attn}
\end{figure}

The proposed model is trained on the dataset and reconstruction profile for 3 weeks is shown in Figure\ref{fig:data3_reco}. Attention maps are first analysed and key observations are noted below:
\label{sec:results3}
\begin{enumerate}
    \item \label{enum:obs1} Figure\ref{fig:data3_feat_attn} shows that the model gives more weightage to the values of same feature across the time for most of the features except f2\_z2\_AC\_3. For this particular unit, weightages are given to 2 features across the time steps. This can be confirmed from the profiles shown in Figure\ref{fig:data3_feat7} and Figure\ref{fig:data3_feat8} where the profiles are closely related since they belong to the same zone. So if one of them is on/off then the reconstruction happens in a way that both or on/off.
    \item \label{enum:obs2} Figure\ref{fig:data3_time_attn} shows that model understood 2 different kinds of periodicity in the data (daily and weekly) seen as horizontal bands. This can be confirmed from the profiles shown in Figure\ref{fig:data3_feat1} and Figure\ref{fig:data3_feat5}.
    \item \label{enum:obs3} Figure\ref{fig:data3_time_attn} shows vertical bands with varying thickness indicating differences in the weightages given to observations on based at a given time of day. The thinner bands on week days (Monday-Saturday) indicate few hours of the day are preferentially weighted to reconstruct windows at all other times of the day. The thicker bands indicate weekend (Sunday) indicate that all times of the given day are identical.
\end{enumerate}

Now, we demonstrate how the above observations help in not only identifying an anomaly but also establishing qualitative differences between the identified anomalies and understand the reason for the model reconstruction based on the relationships it learnt.

\begin{itemize}
    \item In Figure\ref{fig:data3_recof1}, it can be observed that a spike on first wednesday is not reconstructed. From Observation\ref{enum:obs2}, this particular region belongs to the set with Tuesday-Friday period and the model accordingly reduced the weight for the spike to prevent it's reconstruction. 
    \item In Figure\ref{fig:data3_recof7}, A peak is reconstructed by the model on first Tuesday. This region corresponds to the one with period of one day based on the observations from previous days, the model predicted a high power consumption. Also note that the same has been reconstructed in Figure\ref{fig:data3_recof8} since both are related (observation\ref{enum:obs1}).
\end{itemize}

\begin{figure}[ht]
    \centering
    \begin{subfigure}[b]{\textwidth}
        \includegraphics[width=\textwidth]{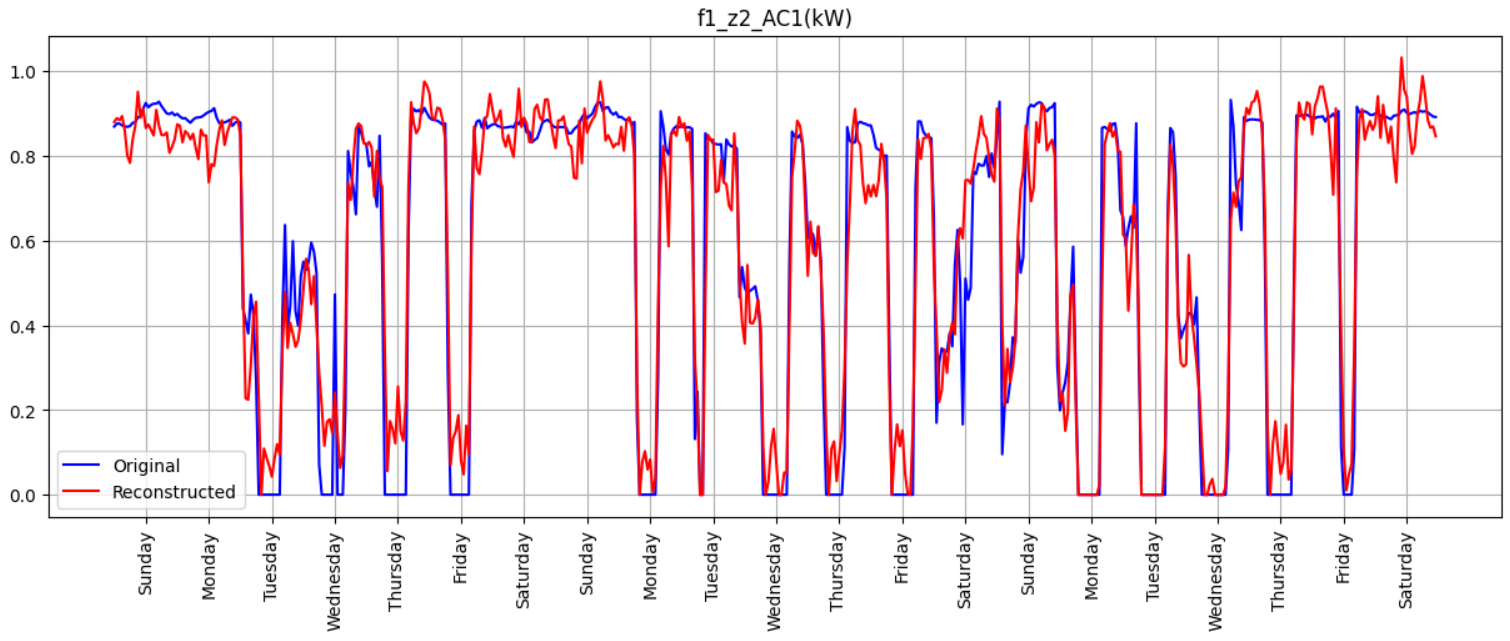}
        \caption{Data-III: Reconstruction vs Original hourly power consumption of Floor 1, Zone 2, AC 1}
        \label{fig:data3_recof1}
    \end{subfigure}
    \begin{subfigure}[b]{\textwidth}
        \includegraphics[width=\textwidth]{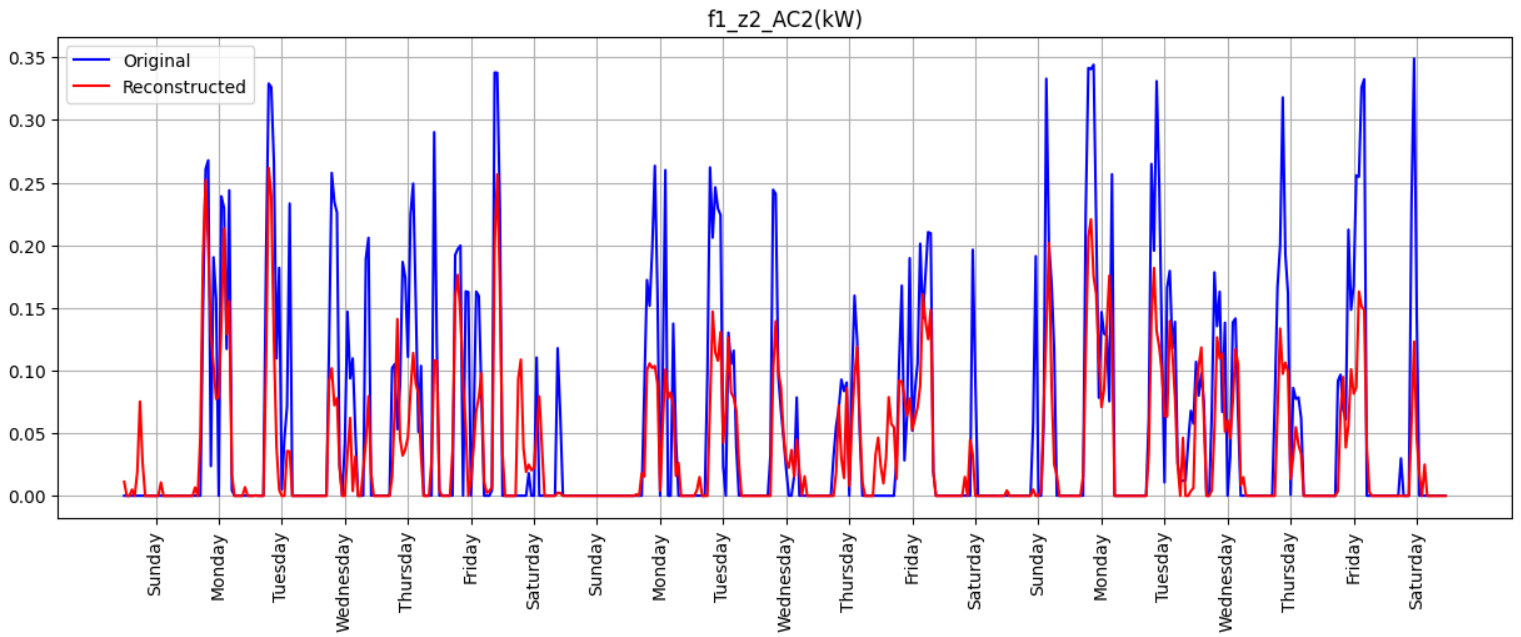}
        \caption{Data-III: Reconstruction vs Original hourly power consumption of Floor 1, Zone 2, AC 2}
        \label{fig:data3_recof2}
    \end{subfigure}
    \begin{subfigure}[b]{\textwidth}
        \includegraphics[width=\textwidth]{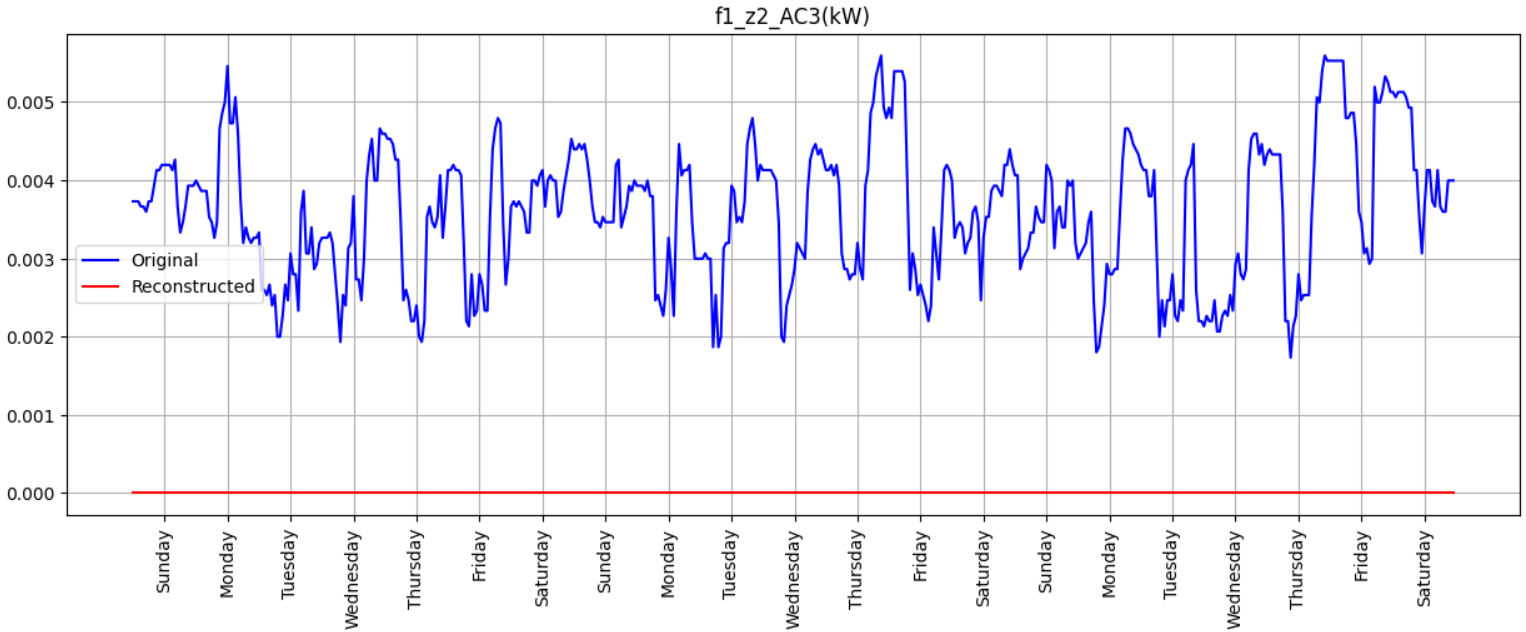}
        \caption{Data-III: Reconstruction vs Original hourly power consumption of Floor 1, Zone 2, AC 3}
        \label{fig:data3_recof3}
    \end{subfigure}
    \caption{Data-III}
\end{figure}
\begin{figure}[ht]\ContinuedFloat
    \centering
    \begin{subfigure}[b]{\textwidth}
        \includegraphics[width=\textwidth]{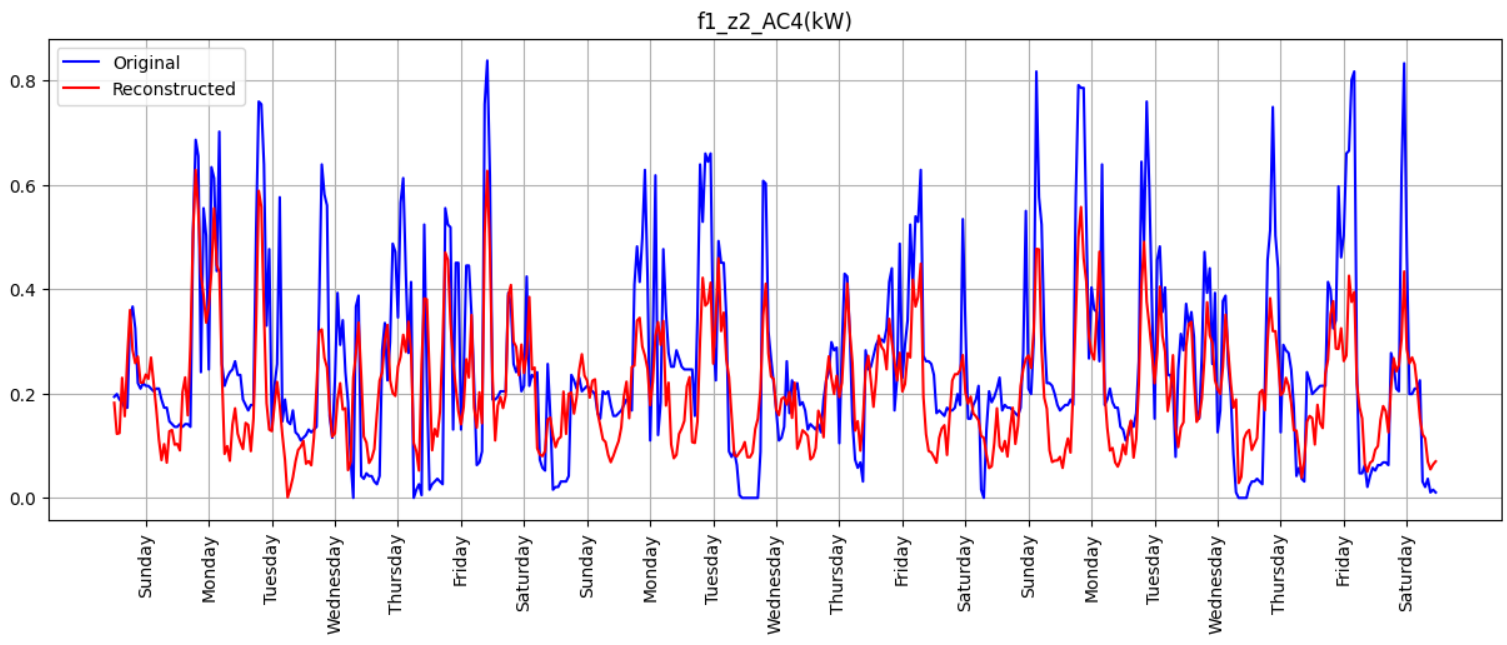}
        \caption{Data-III: Reconstruction vs Original hourly power consumption of Floor 1, Zone 2, AC 4}
        \label{fig:data3_recof4}
    \end{subfigure}
    \begin{subfigure}[b]{\textwidth}
        \includegraphics[width=\textwidth]{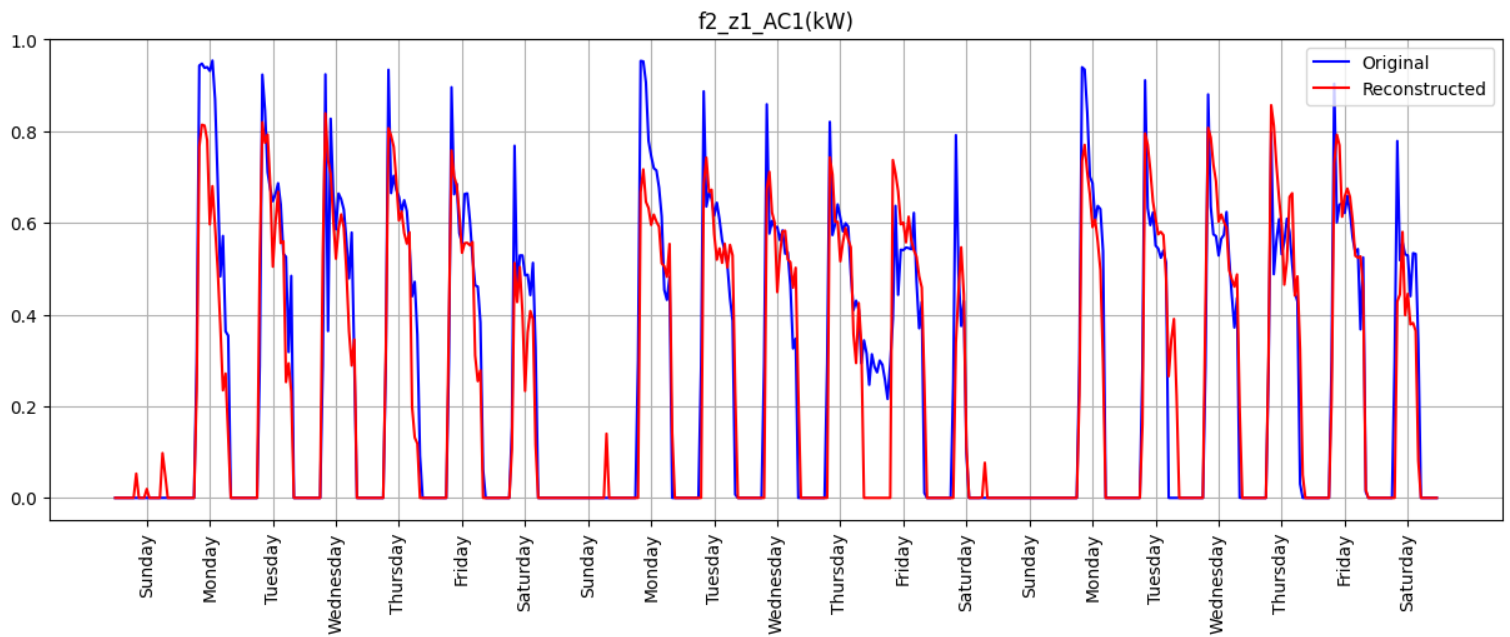}
        \caption{Reconstruction vs Original hourly power consumption of Floor 2, Zone 1, AC 1}
        \label{fig:data3_recof5}
    \end{subfigure}
    \begin{subfigure}[b]{\textwidth}
        \includegraphics[width=\textwidth]{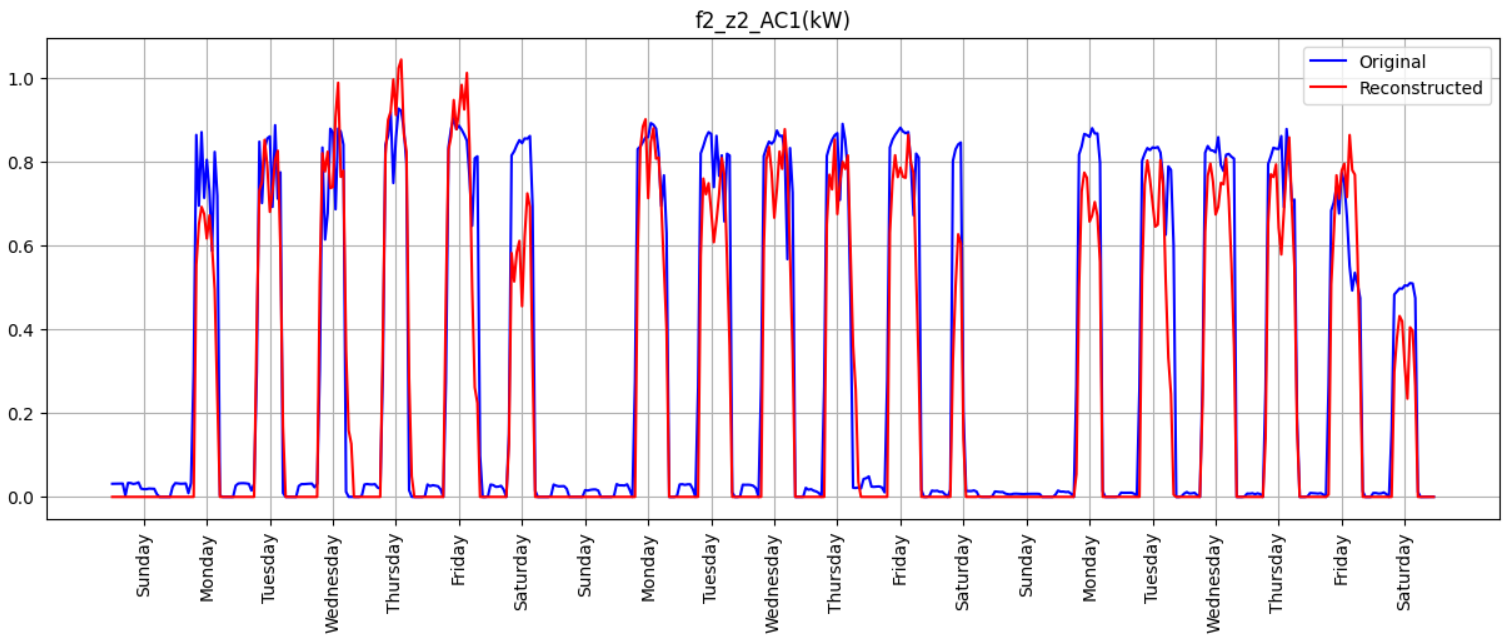}
        \caption{Reconstruction vs Original hourly power consumption of Floor 2, Zone 2, AC 1}
        \label{fig:data3_recof6}
    \end{subfigure}
    \caption{Data-III (contd.)}
\end{figure}
\begin{figure}[ht]\ContinuedFloat
    \begin{subfigure}[b]{\textwidth}
        \includegraphics[width=\textwidth]{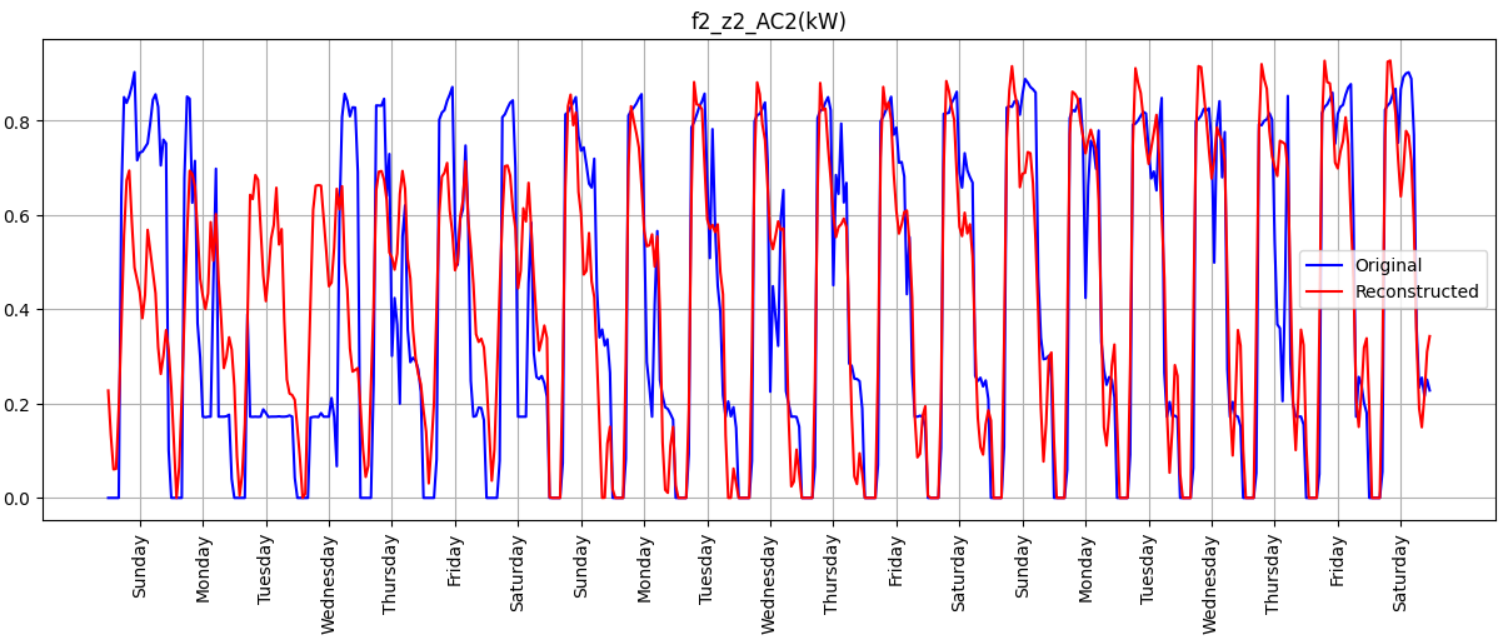}
        \caption{Reconstruction vs Original hourly power consumption of Floor 2, Zone 2, AC 2}
        \label{fig:data3_recof7}
    \end{subfigure}
    \begin{subfigure}[b]{\textwidth}
        \includegraphics[width=\textwidth]{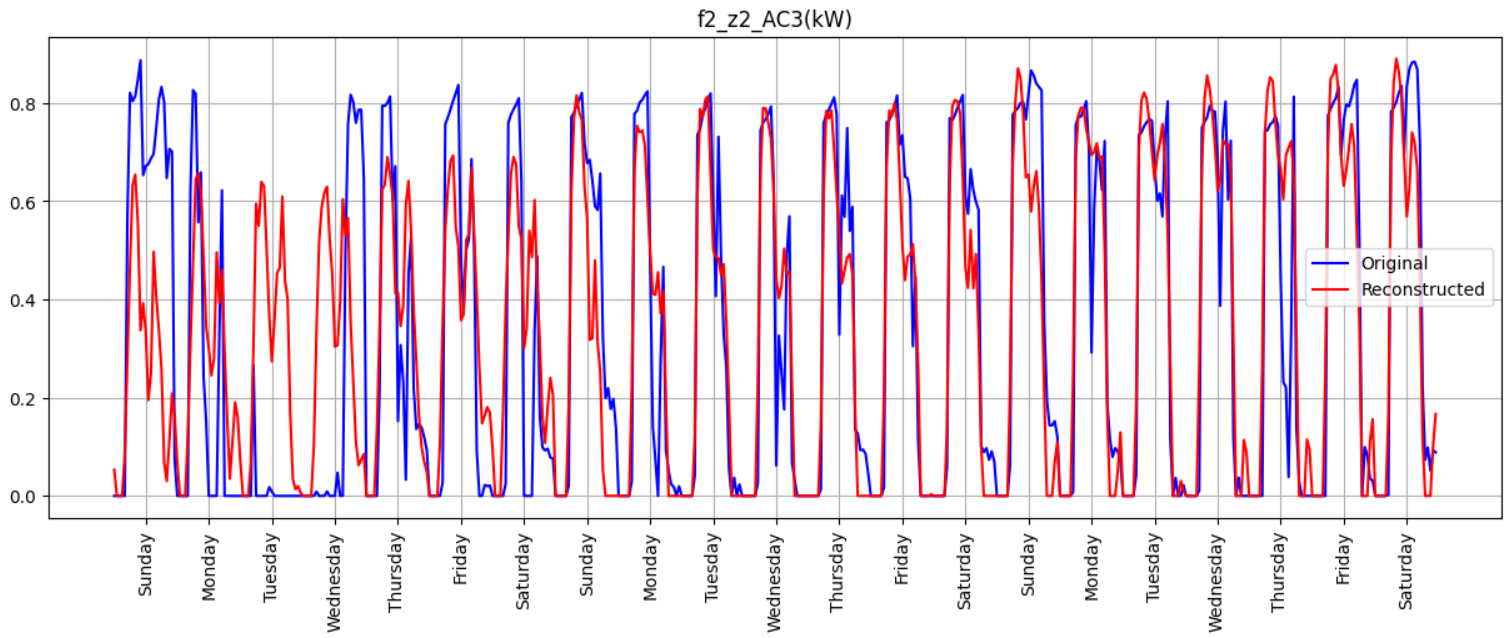}
        \caption{Reconstruction vs Original hourly power consumption of Floor 2, Zone 2, AC 3}
        \label{fig:data3_recof8}
    \end{subfigure}
    \caption{Data-III (contd.)}
    \label{fig:data3_reco}
\end{figure}

\section{Conclusion}
\label{sec:conclusion}
Judicious consumption of energy during building operations can provide significant value towards achieving energy conservation steps. Automated discovery of anomalous consumption patterns can help in developing policies directed to minimise negligent consumer behaviours. The main idea behind the current anomaly detection revolves around identifying normal consumption patterns and raising flags in case of anomalies. Several machine learning techniques are reported to model the normal consumption pattern. The current work proposes a novel attention mechanism to capture the normal consumption behaviour. Sample case-studies demonstrate that the proposed architecture captures these patterns and attention maps thus generated can be used to tune the model parameters without overfitting incase of datasets where training data included both normal and anomalous behaviours (refer Section\ref{sec:results1}. The proposed architecture helps not only in identifying the anomalies but also provides a way to qualitatively classify them using attention maps.

\section{Declaration of competing interest}
The authors declare that they have no known competing financial interests or personal relationships that could have appeared to influence the work reported in this paper.

\section{Data availability}
Synthetically generated data will be made available on request. The real-world dataset is provided in \cite{cubems}.

\section{Code availability}
Code will be made available on request.

 \bibliographystyle{elsarticle-num} 
 \bibliography{cas-refs}

\begin{thebibliography}{10}
\expandafter\ifx\csname url\endcsname\relax
  \def\url#1{\texttt{#1}}\fi
\expandafter\ifx\csname urlprefix\endcsname\relax\def\urlprefix{URL }\fi
\expandafter\ifx\csname href\endcsname\relax
  \def\href#1#2{#2} \def\path#1{#1}\fi

\bibitem{unep_energy}
2022 global status report for buildings and construction: Towards a
  zero‑emission, efficient and resilient buildings and construction sector
  (2022).

\bibitem{iea_report}
\href{https://www.iea.org/reports/buildings}{Iea (2022), buildings, iea,
  paris}.
\newline\urlprefix\url{https://www.iea.org/reports/buildings}

\bibitem{iea_cooling}
\href{https://www.iea.org/reports/the-future-of-cooling}{Iea (2018), the future
  of cooling, iea, paris}.
\newline\urlprefix\url{https://www.iea.org/reports/the-future-of-cooling}

\bibitem{faults}
H.~Rashid, P.~Singh, Monitor: An abnormality detection approach in buildings
  energy consumption, in: 2018 IEEE 4th International Conference on
  Collaboration and Internet Computing (CIC), 2018, pp. 16--25.
\newblock \href {https://doi.org/10.1109/CIC.2018.00-44}
  {\path{doi:10.1109/CIC.2018.00-44}}.

\bibitem{HIMEUR2021116601}
Y.~Himeur, K.~Ghanem, A.~Alsalemi, F.~Bensaali, A.~Amira,
  \href{https://www.sciencedirect.com/science/article/pii/S0306261921001409}{Artificial
  intelligence based anomaly detection of energy consumption in buildings: A
  review, current trends and new perspectives}, Applied Energy 287 (2021)
  116601.
\newblock \href
  {https://doi.org/https://doi.org/10.1016/j.apenergy.2021.116601}
  {\path{doi:https://doi.org/10.1016/j.apenergy.2021.116601}}.
\newline\urlprefix\url{https://www.sciencedirect.com/science/article/pii/S0306261921001409}

\bibitem{vae}
D.~P. Kingma, M.~Welling, \href{https://arxiv.org/abs/1312.6114}{Auto-encoding
  variational bayes} (2013).
\newblock \href {https://doi.org/10.48550/ARXIV.1312.6114}
  {\path{doi:10.48550/ARXIV.1312.6114}}.
\newline\urlprefix\url{https://arxiv.org/abs/1312.6114}

\bibitem{lstm}
S.~Hochreiter, J.~Schmidhuber, Long short-term memory, Neural computation 9~(8)
  (1997) 1735--1780.

\bibitem{vae-lstm}
S.~Lin, R.~Clark, R.~Birke, S.~Schönborn, N.~Trigoni, S.~Roberts, Anomaly
  detection for time series using vae-lstm hybrid model, in: ICASSP 2020 - 2020
  IEEE International Conference on Acoustics, Speech and Signal Processing
  (ICASSP), 2020, pp. 4322--4326.
\newblock \href {https://doi.org/10.1109/ICASSP40776.2020.9053558}
  {\path{doi:10.1109/ICASSP40776.2020.9053558}}.

\bibitem{transformer}
A.~Vaswani, N.~Shazeer, N.~Parmar, J.~Uszkoreit, L.~Jones, A.~N. Gomez,
  L.~Kaiser, I.~Polosukhin, \href{https://arxiv.org/abs/1706.03762}{Attention
  is all you need} (2017).
\newblock \href {https://doi.org/10.48550/ARXIV.1706.03762}
  {\path{doi:10.48550/ARXIV.1706.03762}}.
\newline\urlprefix\url{https://arxiv.org/abs/1706.03762}

\bibitem{transanomaly}
H.~Zhang, Y.~Xia, T.~Yan, G.~Liu, Unsupervised anomaly detection in
  multivariate time series through transformer-based variational autoencoder,
  in: 2021 33rd Chinese Control and Decision Conference (CCDC), 2021, pp.
  281--286.
\newblock \href {https://doi.org/10.1109/CCDC52312.2021.9601669}
  {\path{doi:10.1109/CCDC52312.2021.9601669}}.

\bibitem{cubems}
M.~Pipattanasomporn, G.~Chitalia, J.~Songsiri, C.~Aswakul, W.~Pora,
  S.~Suwankawin, K.~Audomvongseree, N.~Hoonchareon,
  \href{https://doi.org/10.1038/s41597-020-00582-3}{Cu-bems, smart building
  electricity consumption and indoor environmental sensor datasets}, Scientific
  Data 7~(1) (2020) 241.
\newblock \href {https://doi.org/10.1038/s41597-020-00582-3}
  {\path{doi:10.1038/s41597-020-00582-3}}.
\newline\urlprefix\url{https://doi.org/10.1038/s41597-020-00582-3}

\end{thebibliography}





\end{document}